\documentclass[10pt,journal,compsoc]{IEEEtran}
\usepackage[colorlinks, linkcolor=blue, anchorcolor=blue, citecolor=blue]{hyperref} 

\usepackage{doi}
\usepackage{multirow}
\usepackage{tabularx}
\usepackage{booktabs}
\usepackage{subfigure}
\usepackage{xcolor}
\usepackage{comment}
\usepackage{color}
\usepackage{amsmath,amssymb,amsfonts}
\usepackage{algorithmic}
\usepackage{graphicx}
\usepackage{textcomp}
\usepackage{xcolor}
\usepackage{amsmath}
\usepackage{dsfont}
\usepackage[ruled,vlined,linesnumbered]{algorithm2e}
\usepackage{soul}
\newtheorem{theorem}{Theorem}
\SetKwInput{KwInput}{Input}                % Set the Input
\SetKwInput{KwOutput}{Output}              % set the Output
\SetKwInput{KwModel}{Model}                 % set the
\usepackage{xcolor}
\usepackage[normalem]{ulem}
\newcommand{\del}[1]{\sout{\textcolor{blue}{#1}}}
\newcommand{\nt}[1]{\textcolor{red}{#1}}
\usepackage{soul}

\ifCLASSOPTIONcompsoc
  \usepackage[nocompress]{cite}
\else
  \usepackage{cite}
\fi
\ifCLASSINFOpdf
\else
\fi
\begin{document}
\title{Fraud’s Bargain Attack: Generating Adversarial Text Samples via Word Manipulation Process}

\author{Mingze~Ni,        
    Zhensu~Sun,
        and~Wei~Liu~\IEEEmembership{\textit{Senior Member,~IEEE}},
%\thanks{W.Liu is the corresponding author}
\IEEEcompsocitemizethanks{\IEEEcompsocthanksitem Mingze Ni and Wei Liu are with the School of Computer Science, University of Technology Sydney, 15 Broadway, Sydney, 2007, NSW, Australia, \protect e-mail: Mingze.Ni@student.uts.edu.au, Wei.Liu@uts.edu.au. Corresponding author: Wei Liu.
\IEEEcompsocthanksitem Zhensu Sun is with the School of Information Science and Technology, ShanghaiTech
University, 393 Middle Huaxia Road, Shanghai, 201210, China, email: sunzhs@shanghaitech.edu.cn}% <-this % stops an unwanted space
}


\markboth{Journal of \LaTeX\ Class Files,~Vol.~14, No.~8, August~2015}%
{Shell \MakeLowercase{\textit{et al.}}: Bare Demo of IEEEtran.cls for Computer Society Journals}



\IEEEtitleabstractindextext{%
\begin{abstract}
Recent research has revealed that natural language processing (NLP) models are vulnerable to adversarial examples. However, the current techniques for generating such examples rely on deterministic heuristic rules, which fail to produce optimal adversarial examples. In response, this study proposes a new method called the Fraud's Bargain Attack (FBA), which uses a randomization mechanism to expand the search space and produce high-quality adversarial examples with a higher probability of success. FBA uses the Metropolis-Hasting sampler, a type of Markov Chain Monte Carlo sampler, to improve the selection of adversarial examples from all candidates generated by a customized stochastic process called the Word Manipulation Process (WMP). The WMP method modifies individual words in a contextually-aware manner through insertion, removal, or substitution. Through extensive experiments, this study demonstrates that FBA outperforms other methods in terms of attack success rate, imperceptibility and sentence quality.
\end{abstract}

}

\maketitle

\IEEEdisplaynontitleabstractindextext

% \fi
\IEEEpeerreviewmaketitle

\IEEEraisesectionheading{\section{Introduction}\label{introduction}}

AI models that process text data have been widely used in various real-world scenarios, but they are surprisingly vulnerable. Attackers can significantly reduce the performance of these models by intentionally introducing maliciously designed typos, words, and letters into the input \cite{yang2021bigram}. These modified inputs are referred to as adversarial examples. The growing prevalence of adversarial examples raises serious concerns about the comprehension and security of AI models.
\par
A textual attack on an NLP model involves generating adversarial perturbations, which can be classified into character-level, word-level, and sentence-level attacks based on the attack strategy \cite{wang2019towards}. Character-level attacks involve manipulating characters to deceive the model, but this approach can result in misspelled words that can be easily detected and corrected by spell checkers \cite{ebrahimi2017hotflip}. Sentence-level attacks, on the other hand, involve inserting new sentences, paraphrasing sentence pieces, or changing the structure of the original sentences, but they often produce incomprehensible sentences \cite{gan2019improving,hun2021gan}. Therefore, word-level attacks have received more attention from researchers because they are more effective and imperceptible than character-level and sentence-level attacks \cite{wang2019towards}. Attackers typically replace the original word with another heuristically chosen word, but existing methods focus mainly on substitutions, with limited exploration of insertion and removal of words \cite{garg2020bae,zang2020word,li2020bertattack,ni2022attacking}. Furthermore, there is no prior work on using a combination of word substitution, insertion, and removal strategies to generate attacks, which limits the search space for adversarial examples and hinders attack performance. Most algorithms also use a word importance rank (WIR) to launch attacks based on a preset number of perturbed words (NPW) from the top of WIR. However, the optimal NPW varies with different target texts, making presetting challenging \cite{ren2019pwws,li2020bertattack}. Additionally, the different ways of defining WIR, such as probability-weighted word saliency (PWWS) ranking and ranks for drops of classifiers' predictions and transformer logits, make WIR-based methods ineffective in achieving imperceptibility, as they may change the semantics of the original sentences.

\par

This study proposes a new attacking algorithm, called Frauds' Bargain attack (FBA), to tackle the limitations of existing word-level attacks. FBA utilizes a stochastic process called Word Manipulation Process (WMP), which considers word substitution, insertion, and removal strategies to generate potential adversarial candidates. The Metropolis-Hasting algorithm is used to select the best candidates based on a customized acceptance probability, which minimizes semantic deviation from the original sentences. WMP proposes malicious candidates that could be potential adversarial examples, while MH algorithm acts as an experienced 'fraud' to statistically evaluate their quality and select the best candidates.

\par
FBA has two key advantages over most existing attacks in the literature. Firstly, FBA offers a large search space for generating adversarial examples by allowing word manipulation through insertion, substitution, and deletion, unlike many attacks that rely solely on word substitutions. This increased search space allows for a theoretically higher probability of generating effective adversarial examples. Secondly, FBA employs a customized word distribution to stochastically select the attacked positions, enabling each word in the context to be chosen based on its importance. This approach allows for greater flexibility and adaptability compared to methods that rely on presetting the number of perturbed words (NPW). Additionally, FBA regulates the Metropolis-Hasting (MH) sampler based on the imperceptibility of the attacks and minimal deviation from the original semantics, resulting in an adaptive setting of NPW that can vary depending on the input text. This approach allows for more successful attacks, as the optimal NPW can differ depending on the characteristics of the input text.

\par
In summary, our work addresses the challenge of generating effective and semantically preserving adversarial examples for natural language processing tasks, and the main contributions in this work are as follows:
\begin{itemize}
    \item We develop a probabilistic approach called the Word Manipulation Process (WMP) to generate a wide range of potential adversarial examples. The WMP involves the manipulation of words through operations such as insertion, removal, and substitution, which provides a large search space for generating adversarial candidates.
    \item Our paper introduces a powerful adversarial attack method called Fraud's Bargain Attack (FBA) that uses the Metropolis-Hasting (MH) algorithm to construct an acceptance probability. This probability is then applied to select the highest-quality adversarial candidates generated by the Word Manipulation Process (WMP). By using the acceptance probability, our attack method can avoid getting stuck in local optima and can generate solutions that are closer to the global optima.
    \item In order to demonstrate the effectiveness of our proposed Fraud's Bargain Attack (FBA) method, we conducted experiments on real-world public datasets. Our experimental results indicate that our method outperformed other state-of-the-art methods in terms of attack success rate, preservation of semantics, and overall sentence quality. These findings provide strong evidence to support the effectiveness of our proposed method in generating high-quality adversarial examples while preserving the original semantics of the input sentence.

\end{itemize}

The rest of this paper is structured as follows. We first review adversarial attacks for NLP models and the applications of Markov Chain Monte Carlo (MCMC) methods to NLP models in Section \ref{related work}. We detail our proposed method in Section \ref{WMP} and \ref{fba} and then evaluate the performance of the proposed method through empirical analysis in Section \ref{experiments}. We conclude the paper with suggestions for future work in Section \ref{conclusion}.
\section{Related Work}\label{related work}
In this section, we provide a review of the literature on word-level textual attacks and MCMC sampling in NLP. 
\subsection{Word-level Attacks to Classifiers}
Word-level attacks pose non-trivial threats to NLP models by locating important words and manipulating them for targeted or untargeted purposes. With the help of an adopted fast gradient sign method (FGSM) \cite{Goodfellow2015fgsm}, Papernot \cite{PapernotMSH16} was the first to generate word-level adversarial examples for classifiers. Notably, while textual data is naturally discrete and more perceptible than image data, many gradient-based textual attacking methods inherited from computer vision are not effective enough, making textual attack a challenging problem to address. Alzantot \cite{alzantot2018generating} transferred the domain of adversarial attacks to an optimization problem by formulating a customized objective function. With genetic optimization, they succeeded in generating adversarial examples by sampling the qualified genetic generations that break out of the encirclement of the semantic threshold. However, genetic algorithms can be inefficient. Since the word embedding space is sparse, performing natural selection for languages in such a space can be computationally expensive. Jia \cite{jia2019faga} proposed a faster version of Alzantot's adversarial attacks by shrinking the search space, which accelerates the process of evolving in genetic optimization. Although Jia has greatly reduced the computational expense of genetic-based optimization algorithms, searching or optimizing algorithms in word embedding spaces, such as GloVe \cite{pennington2014glove} and Word2Vec \cite{Mikolov2013word2vect}, are still not efficient enough.

To ease the searching process, embedding-based algorithms often resort to using a counter-fitting method to post-process attacker’s vectors to ensure the semantic constraint \cite{mrkvsic2016counter}. Compared with word embedding methods, utilizing a well-organized linguistic thesaurus, e.g., synonym-based WordNet \cite{miller1990wordnet} and sememe-based HowNet \cite{dong2003hownet}, offers a simpler and more straightforward implementation. Ren \cite{ren2019pwws} sought synonyms based on WordNet synsets and ranked word replacement order via probability-weighted word saliency (PWWS). Zang \cite{zang2020word} and Yang \cite{yang2021bigram} both demonstrated that the sememe-based HowNet can provide more substitute words using the Particle Swarm Optimization (PSO) and an adaptive monotonic heuristic search to determine which group of words should be attacked. Additionally, some recent studies utilized a masked language model (MLM), such as BERT \cite{Devlin2019BERTPO} and RoBERTa \cite{Liu2019RoBERTaAR}, to generate contextual perturbations \cite{li2020bertattack,garg2020bae}. The pre-trained MLM ensures that the predicted token fits the sentence's grammar correctly but might not preserve the semantics.

\subsection{Markov Chain Monte Carlo in NLP}
Markov Chain Monte Carlo (MCMC) stands as a versatile technique for approximating samples from complex distributions, finding critical applications in areas like optimization \cite{rubinstein1999crossoptimization}, machine learning \cite{fan2018rectangular}, and quantum simulation \cite{haase2021quantum}. Among the myriad MCMC algorithms, the Metropolis-Hastings (MH) algorithm is arguably the most foundational. Tracing its roots to the pioneering work of Metropolis et al. \cite{metropolis1953MH}, and later extended by Hastings \cite{hastings1970monte}, the MH algorithm emerges as an indispensable tool in the MCMC arsenal. Its primary utility lies in generating a sequence of samples from a target distribution, especially when direct sampling poses challenges. The algorithm achieves this by formulating a Markov chain whose equilibrium distribution coincides with the desired distribution. At its core, the MH algorithm utilizes a proposal distribution to propose new candidate samples and employs an acceptance criterion to assure convergence to the intended distribution. While the MH algorithm's adaptability makes it suitable for a broad spectrum of applications, its efficiency hinges on the meticulous design of the proposal distribution. In the field of Natural Language Processing (NLP), MCMC methods, including MH, have been instrumental in tasks such as text generation, sentiment analysis, and named entity recognition, aiding in the evolution of models that interpret and comprehend human language.

\par
For text generation, Kumagai \cite{Kumagai2016HumanlikeNL} propose a probabilistic text generation model which generates human-like text by inputting semantic syntax and some situational content. Since human-like text requests grammarly correct word alignment, they employed Monte Carlo Tree Search to optimize the structure of the generated text. In addition, Harrison \cite{harrison2017toward} presents the application of MCMC for generating a story, in which a summary of movies is produced by applying recurrent neural networks (RNNs) to summarize events and directing the MCMC search toward creating stories that satisfy genre expectations. For sentimental analysis, Kang \cite{Kang2011SamplingLE} applies the Gibbs sampler to the Bayesian network, a network of connected hidden neurons under prior beliefs, to extract the latent emotions. Specifically, they apply the Hidden Markov models to a hierarchical Bayesian network and embed the emotions variables as the latent variable of the Hidden Markov model. To accelerate the training speed of name entity recognition models, Sameer \cite{Singh2012MonteCM} proposes an adaptive MCMC sampling scheme for a graphical NLP model. They demonstrate that the proposed sampler can sample a set of latent variables which are asymptotic to the original transition distribution. The proposed sampler converges more quickly than gradient-based optimization from their experiments, and the sampled latent variables can represent the critical name entities.

\section{Word Manipulation Process}\label{WMP}
\noindent  In this section, we provide a detailed explanation of the Word Manipulation Process (WMP), which is used to generate adversarial candidates. The method for selecting the generated candidates will be described in the following section. Let $D={(x_1,y_1),(x_2,y_2),\ldots,(x_m,y_m)}$ be a dataset consisting of $m$ samples, where $x$ and $y$ represent the input text and its corresponding class, respectively. The victim classifier $F$ maps from the text space $\mathcal{X}$ to the class space using a categorical distribution, such that $F(\cdot):\mathcal{X}\rightarrow{(0,1)^K}$, where $K$ is the number of classes. Given an input text $x$ consisting of $n$ words $w_1, w_2, ..., w_n$, we refer to a modified version of the text as an adversarial candidate and denote it as $x^\prime$. The final chosen adversarial example is referred to as $x^*$.

\par
The Word Manipulation Process (WMP) takes three steps to modify the current text state $x$. Firstly, an action $e$ is randomly selected from the set of possible actions, which includes inserting a word at the beginning of the sentence, substituting a word, or removing a word. Secondly, the position $l$ where the manipulation should take place is determined by drawing from a customized categorical distribution. Thirdly, if word insertion or substitution is selected, WMP generates a candidate by combining synonyms of the original words and using a pre-trained masked language model (MLM) such as BERT \cite{Devlin2019BERTPO}, XML \cite{Conneau2020UnsupervisedCR} and MPNet \cite{song2020mpnet}. The algorithm for WMP is presented in Algorithm \ref{algo:WMP}. In the following, we elaborate on the details of the three steps of WMP:

\begin{algorithm}[t]
\caption{Word Manipulation Process}
\label{algo:WMP}
\DontPrintSemicolon
\KwInput{Number of iterations: $T$, Input text: $x$}
\KwOutput{A set of adversarial candidates: $[x_{1}^{\prime},\ldots,x_{T}^{\prime}]$}
candidates=[ ]\;
$x_{0}^{\prime}=x$
\For{$t$ in $T$}{
Given $x_{t-1}^{\prime}$, sample $e$ with Eq. \ref{eqt: action}\;
Given $x_{t-1}^{\prime}$, $e$, sample $l$ with Eq. \ref{eqt: position dis}\;
Given $x_{t-1}^{\prime}$, $e$, $l$, sample $o$ with Eq. \ref{eqt: candidate dis}\;
Craft $x_t^{\prime}$ by taking action $e,l,o$ to $x_{t-1}^{\prime}$\;
candidate.append($x_{t}^{\prime}$)}
\Return candidates
\end{algorithm}

\subsection{Action}
We draw $e \in \{0,1,2\}$ from a categorical distribution: 
\begin{align}
p(e\vert x)& = 
    \begin{cases}
        \mathbf{P}_{ins}& e=0 ,\\
        \mathbf{P}_{sub} & e=1 ,\\
       \mathbf{P}_{rem} & e=2\\
    \end{cases}\\ \label{eqt: action}
   \nonumber
\end{align}
where $\mathbf{P}_{ins}+\mathbf{P}_{sub}+\mathbf{P}_{rem}=1$, and $\mathbf{P}_{ins}, \mathbf{P}_{sub},$ and $\mathbf{P}_{rem}$ represent probability of insertion ($0$), substitution ($1$) and removing ($2$), respectively. %In general, we set $\mathbf{P}_{ins}=\mathbf{P}_{rem}$, as we believe removal and insertion are equally important. 
The probabilities of these types of attacks can be set by the attacker's preference.

\subsection{Position}
Given a certain action $e$, we need to select one target word at location $l$ in the sentence to implement the attack. Considering the effectiveness of the selection, higher probabilities should be assigned to the words with more influence. To solve this, we use the changes of victim classifiers’ logits, $I=[I_{w_1},\ldots,I_{w_i},\ldots,I_{w_n}]$, before and after deleting a word. Such a drop of logits for $i$th word, $I_{w_i}$, is mathematically formulated as:
\begin{align}
    I_{w_i}=F_{y,logit}(x)-F_{y,logit}(x_{w_i}),
\end{align}
where $F_{y,logit}(\cdot)$ is the classifier returning the logit of the correct class, and $x_{w_i}=[w_1,\ldots ,w_{i-1},w_{i+1},\ldots, w_{n}]$ is the text with $w_i$ removed. Different from word importance rank (WIR), we utilize drops of logits $I$ to craft categorical distribution on position $l$, $p(l\vert e, x)$, by putting $I$ to a softmax function as follows:
\begin{align}
    p(l\vert e, x)=\mathrm{softmax}(I) \label{eqt: position dis}
\end{align}
This way, locations of words (tokens) are assigned probabilities according to the words' influence on the classifier.

\subsection{Word Candidates}
Different actions require different search strategies for word candidates. 
% For different actions, we design different search strategies for word candidates.
To find the word candidates for substitution and insertion, we utilize an MLM and synonyms of the original words (calculated by nearest neighbours using the L2-norm of word embeddings), for parsing fluency and semantic preservation, respectively. As for word removals, we design a hesitation mechanism to maintain a probabilistic balance with word insertion. The details are demonstrated in the following paragraphs.
\par
\subsubsection{Candidates for Substitution Attacks}
We mask the word on the selected position to construct 
\begin{align*}
x_{sub}^*=[w_{1},\ldots,[MASK],\ldots,w_{n}]
\end{align*}
and feed this masked sentence $x_{sub}^*$ into a MLM $\mathcal{M}(\cdot):\mathcal{X}\rightarrow{(0,1)^d}$ to obtain a distribution about word candidates $o$ across all the words in the dictionary size $d$. The distribution is below:
\begin{align}
    p_{\mathcal{M}}(o\vert l, e, x)=\mathcal{M}(x^*_{sub}) \label{eqt: mlm}
\end{align}

The MLM relies on softmax to output a distribution on the dictionary, but most words from the dictionary can be grammarly improper, and the probability of selecting one of these words can be high. To this end, we tend to create another distribution with respect to the $k$ top word candidates from the MLM. By mixing such a distribution with the MLM distribution, the probability of selecting grammarly improper words can be decreased. To construct such a $k$ top words distribution, we choose the $k$ top words $G^{sub}_{\mathcal{M}}=\{w^{sub}_{(\mathcal{M},1)},\ldots,w^{sub}_{(\mathcal{M},k)}\}$ from $\mathcal{M}$ and treat every word from this set equally important:
\begin{align}
    p^{top}_{\mathcal{M}}(0\vert l, e, x)=\mathds{1}(o \in G^{sub}_{\mathcal{M}})\frac{1}{k}, \label{eqt: mlm top}
\end{align}
where $\mathds{1}(\cdot)$ is an indicator function. In such a setting, the top candidates from MLM are attached with more importance.
\par
Synonyms often fit more seamlessly with parsers in MLM, and they generally achieve a higher probability in MLM, as evidenced by studies such as \cite{li2021clare,Jin2020IsBR}. While MLM has already demonstrated its prowess in maintaining semantics, we emphasize a robust selection of synonyms based on word embeddings. This approach to synonym selection has been corroborated as effective by research like \cite{alzantot2018generating, jia2019faga, Wang2019NaturalLA}. Specifically, we perform synonym extraction by gathering a word candidate set for top $k$ replacements of the selected word. Specifically, we use L-2 norm as the metric to perform kNN inside word embedding space from BERT and construct such a synonym candidates set $G_{nn}=\{w_{(\mathcal{M},1)},\ldots,w_{(\mathcal{M},k)}\}$, with top-k nearest neighbors from the embedded spaces as synonyms for the word on selected position $l$. %. With such a word candidate set, we use the method of crafting $p^{top}_{\mathcal{M}}(o\vert l, e, x)$ in Eq. \ref{eqt: mlm top}
In this way, we construct the synonym words distribution as follows:
\begin{align}
    p_{nn}(0\vert l, e, x)=\mathds{1}(o \in G_{nn})\frac{1}{k} \label{eqt: syn}
\end{align}
As we tend to generate parsing-fluent and semantic-preserving adversarial candidates, we combine the distributions in Eq. \ref{eqt: mlm}, Eq. \ref{eqt: mlm top} and Eq. \ref{eqt: syn} to construct a mixture distribution as the final distribution to draw the substitution:
\begin{align}
    p_{sub}(o\vert e,l,x)=& a_{1}p_{\mathcal{M}}(o\vert l, e, x)+a_{2}p^{top}_{\mathcal{M}}(o\vert l, e, x) \nonumber\\
    &+a_{3}p_{nn}(o\vert l, e, x) ,\label{eqt: sub dis}
\end{align}
where $a_{1}+a_{2}+a_{3}=1$, $a_{1},a_{2},a_{3}\in (0,1)$, while $a_{1}$, $a_{2}$ and $a_{3}$ are hyper-parameters for weighing the corresponding distribution.
\subsubsection{Candidates for Insertion Attacks}
Searching word candidates for insertion attacks follows a similar logic as substitutions but without a synonym search. We construct the mask sentence $x_{ins}^*=[w_{1},\ldots,w_{l-1},[MASK],w_{l},\ldots,w_{n}]$ by inserting a masked token on the left side of a selected position, then apply the MLM $\mathcal{M}$ to the mask sentence for extracting the output of the softmax layer, $\mathcal{M}(x_{ins}^*)$. Following the same logic as Eq. \ref{eqt: mlm top}, we select the top $k$ word candidates $G^{ins}_{\mathcal{M}}=\{w_{(\mathcal{M},1)},\ldots,w_{(\mathcal{M},k)}\}$. Similar to Eq. \ref{eqt: sub dis}, insertion word candidate distribution can be constructed as follows:
\begin{align}
    p_{ins}(o\vert l,e,x)&=b_{1}\mathcal{M}(x_{ins})+b_{2}\mathds{1}(o \in G^{ins}_{\mathcal{M}})\frac{1}{k} \label{eqt:insert}
\end{align}
where  
$b_{1}+b_{2}=1$, $b_{1},b_{2}\in (0,1)$, $b_{1}$ and $b_{2}$ are hyper-parameters for weighing the corresponding distribution. 
\subsubsection{Candidates for Removal Attacks} %One could remove the selected word directly without searching for word candidates. However, in practise the same word may be removed multiple times on one sentence across different stochastic runs, resulting in the same adversarial candidates. To avoid this unideal scenario, w
Since word candidates for insertion and substitution can be drawn from a significantly large dictionary, these two actions will provide a large variance of adversarial candidates. Differently, removal does not require a selection of word candidates but directly removes the word on the selected position, which will lead to a low variety of adversarial candidates crafted by removal. The consequence of such a low variety is that the probability of crafting the same adversarial candidate is much higher than inserting and substituting. To balance such a probability, we design a Bernoulli distribution for determining word removals. Specifically, we craft the removal word candidates set $G^{rem}=\{0,1\}$, where 0 and 1 represent remaining and removing the selected word, respectively. The distribution is as follows:%, and the empty set $\emptyset$ is chosen with probability $\frac{1}{k}$:
\begin{align}
    p_{rem}(o\vert e,l, x) = &
    \begin{cases}
        1-\frac{1}{k} & o=0,\\
        \frac{1}{k}& o=1
        %,\\
        %0& o \notin G^{rem}
    \end{cases}\label{eqt: remove dis}
\end{align}
With the above distribution, the $o=1$ (i.e., to remove the word) is selected to replace the original word with probability $\frac{1}{k}$. The rationale of using $\frac{1}{k}$ is to decrease the probability of repeatedly proposing the same perturbed sentence with action removal such that it is approximate to the probability of word replacement and insertion as in Eq.~\ref{eqt: mlm} and Eq.~\ref{eqt:insert}: while each replacement word has a probability of $\frac{1}{k}$ for being chosen, the removal, $o=1$, has the same probability of being selected in removal attacks.

With the three WMP steps, we summarize the probability density function for word candidates:% in Eq. \ref{eqt: candidate dis}.
\begin{align}
    p(o\vert e,l, x)=&
    \begin{cases}
        p_{ins}(o\vert e,l, x)& e=0 ,\\
        p_{sub}(o\vert e,l, x)& e=1,\\
        p_{rem}(o\vert e,l, x)& e=2
    \end{cases}\label{eqt: candidate dis}
\end{align}
\par
\subsection{Integration of the three WMP steps}
%By carrying WMP one time with the three-step manipulation, an adversarial candidate $x^\prime$ can be constructed with conditional sampling $e$, $l\vert e$ and $o\vert l,e$ given the input text $x$.
By iteratively running WMP $T$ times with an initial start at original input text ($x^{\prime}_{0}=x$), we can get a sequence of adversarial candidates $x^{\prime}_{T}=[x^{\prime}_{1},
\ldots,x^{\prime}_{t},\ldots,x^{\prime}_{T}]$.
By applying the Bayes rule, we can derive the WMP's distribution from the iteration $t$ to $t+1$ as the following equation:
\begin{align}
    \mathrm{WMP}(x^{\prime}_{t+1}\vert x^{\prime}_{t})=&p(e,l,o\vert x^{\prime}_{t}) \nonumber\\
    =&p(e\vert x^{\prime}_{t})p(l\vert e, x^{\prime}_{t})p(o\vert e, l, x^{\prime}_{t})\label{eqt: final proposing function}
\end{align}

\noindent\subsubsection*{\textbf{Two Theoretical Merits of WMP}} 
WMP is expected to own two major merits: enlarging the searching domain and the ability to correct possible wrong manipulation. These two merits can be guaranteed by the aperiodicity shown in the following theorem.
\begin{theorem}
Word Manipulation Process (WMP) is aperiodic. \label{theo: WMP reverse}
\end{theorem}
% The proof of Theorem \ref{theo: WMP reverse} is below:
\begin{IEEEproof}
Suppose we have two arbitrary text samples $x_{i},x_{j} \in \mathcal{X}$ from text space $\mathcal{X}$. $x_{i}=[w_{1}^{i},\ldots,w^{i}_{n_{i}}]$ $x_{j}=[w_{1}^{j},\ldots,w_{1}^{n_{j}}]$ have $n_{i}$ and $n_{j}$ words, respectively. To prove the process is aperiodic by definition, we need to show: 
\begin{align}
    \exists N \leq \infty \text{,~} \;\mathbb{P}(x^{(N)}=x_{j}\vert x_{i})>0,
\end{align}
which means that there always exists $\exists N \leq \infty$ that can make the probability of transferring $x_{i}$ to $x_{j}$, after $N$ steps, greater than zero. % $\mathbb{P}(x^{(N)}=x_{j}\vert x_{i})$. 
\par
Because a text dataset is discrete and WMP is time-discrete, WMP is a Markovian process. Therefore, we apply the Chapman–Kolmogorov equation \cite{pavliotis2015stochastic}, to derive the following equation:
\begin{align}
    \mathbb{P}\left(x^{(N)}=x_{j}\vert x_{i}\right)&=\sum_{x^{(t)}\in \mathcal{X}} \mathrm{WMP}\left(x^{(1)}\vert x_{i}\right)\nonumber\\ 
    \cdots\mathrm{WMP}&\left(x^{(t)}\vert x^{(t-1)}\right)\cdots\mathrm{WMP}\left(x_{j}\vert x^{(N-1)}\right),
\end{align}
where $\mathrm{WMP}(\cdot)$ denotes the Word Manipulation process (WMP) and $x^{(t)}$ denotes the arbitrary text in text space $\mathcal{X}$ after runing WMP for $t$ times. We try to prove aperiodicity with a special case, let $N=n_{i}+n_{j}$ $\mathcal{A}$: first inserting all the words from $x_{j}$ to the $x_{i}$, then remove all words from $x_i$. This process can be illustrated as follows:
\begin{align*}
    \mathcal{A}: \quad x_{i} &\xrightarrow[insert]{n_{j} \mathrm{times}}{} x^{n_{j}}=[w_{1}^{i},\ldots,w^{i}_{n_{i}}, w_{1}^{j},\ldots,w_{1}^{n_{j}}]\\
    &\xrightarrow[remove]{n_{j} \mathrm{times}}{}x_{j}=[w_{1}^{j},\ldots,w_{1}^{n_{j}}]
\end{align*}
As $\mathcal{A}$ is the special case of the $N$ times iterations, we have:
\begin{align}
\mathbb{P}(\mathcal{A})\leq \mathbb{P}\left(x^{(N)}=x_{j}\vert x_{i}\right).
\end{align}
Moreover, the WMP inserts one word on any position based on softmax, which outputs non-zero probabilities, therefore we can derive:
\begin{align}
0<\mathbb{P}(\mathcal{A})\leq \mathbb{P}\left(x^{(N)}=x_{j}\vert x_{i}\right).
\end{align}
Therefore, we find that for arbitrary $x_{i}, x_{j} \in \mathcal{X}$, there always exist $ N=n_{i}+n_{j} \leq \infty$ that can make the probability of transfer $x_{i}$ to $x_{j}$ after $N$ time larger than zero, i.e.,  $\mathbb{P}(x^{(N)}=x_{j}\vert x_{i})>0$. According to the definition, we have successfully proved Theorem 1.
\end{IEEEproof}

\begin{algorithm}[t]
\caption{Fraud's Bargain Attack (FBA)}
\label{algo: FBA}
\DontPrintSemicolon
\KwInput{Input text: $x$, Number of Sample: $T$}
\KwOutput{An adversarial example}
$Adv\_set=[\quad]$\;
$x_{1}=x$\;
\For{ t in range($T$)}{
Sample $x_{t}$ from WMP given $x_{t-1}$ with Eq. \ref{eqt: final proposing function}\;
Sample $u$ from Uniform distribution, Uniform(0,1)\;
Calculate the acceptance probability, $prob=\alpha(x_{t+1},x_{t})$ with Eq. \ref{eqt: accept prob}\;
\uIf{$u<prob$}
{
$x_t=x_{t+1}$\;
$Adv\_set$.append($x_{t+1}$)
}
\Else{
$x_t=x_t$\;
$Adv\_set$.append($x_{t}$)
}
\Return $Adv\_set$
}
Choose the candidate with the least modification as the adversarial example $x^*$.\;
\Return adversarial example $x^*$
\end{algorithm}
\par
Aperiodicity implies that, given a large enough number of iterations $T$, an arbitrary $x$ can be perturbed to any text $x^{\prime}$ in text space, i.e., $\mathrm{WMP}(x\vert x^{\prime})\neq 0$. The first merit of aperiodicity, WMP theoretically guarantees that the searching domain is enlarged to generate the most effective adversarial candidates. Since WMP samples the adversarial candidates with randomness, there is a tiny possibility of crafting a bad adversarial candidate. With aperiodicity, WMP is eligible for correcting this bad manipulation by reversing the bad sample $x^{\prime}_{t+1}$ to the previous state $x^{\prime}_{t}$.

\section{Adversarial Candidate Selection by Metropolis-Hasting Sampling}\label{fba}
\noindent A simple approach to selecting adversarial examples is to test all candidates generated by WMP and choose the best-performing one. However, this method can be time-consuming and may result in over-modified adversarial examples. To address this issue, we propose the Fraud's Bargain Attack (FBA), which enhances WMP using the Metropolis-Hastings (MH) algorithm. FBA selects adversarial candidates by evaluating them using a customized adversarial distribution.

\subsection{Markov Chain Monte Carlo}

Markov chain Monte Carlo (MCMC), a statistically generic method for approximate sampling from an arbitrary distribution, can be applied in a variety of fields.
% , such as optimization \cite{rubinstein1999crossoptimization}, machine learning \cite{fan2018rectangular}, quantum simulation \cite{haase2021quantum} and icing models \cite{herrmann1986ising}.
The main idea is to generate a Markov chain whose equilibrium distribution is equal to the target distribution \cite{kroesehandbook}. Metropolis Hasting (MH) sampler \cite{metropolis1953MH}, from which MCMC methods originate, applies the following setting: suppose that we wish to generate samples from an arbitrary multidimensional probability density function (PDF):
\begin{align}
    f(\mathbf{x})=\frac{p(\mathbf{x})}{\mathcal{Z}}, 
\end{align}
where $p(\mathbf{x})$ is a known positive function and $\mathcal{Z}$ is a known or unknown normalizing constant. Let $q(\mathbf{y} | \mathbf{x})$ be a proposal or instrumental density: a Markov transition density describing how to go from state $\mathbf{x}$ to $\mathbf{y}$. The MH algorithm is based on the following ``trial-and-error" strategy by defining an accept probability $\alpha(\mathbf{y}\vert \mathbf{x})$ as following:
\begin{align}
    \alpha(\mathbf{y}\vert \mathbf{x})=\min \left\{\frac{f(\mathbf{y}) q(\mathbf{x} \mid \mathbf{y})}{f(\mathbf{x}) q(\mathbf{y} \mid \mathbf{x})}, 1\right\}
\end{align}
to decide whether the new state $y$ is accepted or rejected. To be more specific, we sample a random variable $u$ from a Uniform distribution ranging 0 to 1, $u\sim Unif(0,1)$, and if $u<\alpha(\mathbf{x}, \mathbf{y})$ then $\mathbf{y}$ is accepted as the new state, otherwise the chain remains at $\mathbf{x}$. The fact that equilibrium distribution of MH Markov Chain is equal to the target distribution is guaranteed by the local/detailed balance equation.
\par
A good property of MH sampler is that in order to evaluate the accept rate $\alpha(\mathbf{x}, \mathbf{y})$, it is only necessary to know the positive function $q(\mathbf{y} \mid \mathbf{x})$, which is also known as kernel density. In addition, the efficiency of the MH sampler depends on the choice of the transition density function $q(\mathbf{y} \mid \mathbf{x})$. Ideally, $q(\mathbf{y} \mid \mathbf{x})$ should be "close"
to $f(\mathbf{y})$, with respect to $\mathbf{x}$.

\subsection{Adversarial Distribution}
We argue that adversarial examples should work with imperceptible manipulations to input text. Therefore, given text $x$, we construct the adversarial target distribution $\pi(x^\prime):\mathcal{X} \rightarrow{(0,1)}$ to measure the classifier's deprivation once under attack with a heavy penalty on change of semantics. Concretely, we measure the classifier's deprivation by defining a measure of distance to perfection, $R$, based on the confidence of making wrong predictions $1-F_{y}(x^{\prime})$, where $F_{y}:\mathcal{X}\rightarrow{[0,1]}$ is the confidence of predicting correct class. The higher the value of the distance to perfection $R$, the more successful the attack. Meantime, we add a regularizer on semantic similarity, $\mathrm{Sem}(\cdot)$. Thus the mathematical formulation is as follows:
\begin{align}
    \pi(x^{\prime}&\vert x, \lambda)=\frac{R+\lambda \mathrm{Sem}(x^{\prime},x)}{C} \label{eqt: target dis}\\[0.9em]
    R = &
    \begin{cases}
        \displaystyle 1-F_{y}(x^{\prime})& F_{y}(x^{\prime})>\frac{1}{K} ,\\[0.8em]
       \displaystyle 1-\frac{1}{K}& F_{y}(x^{\prime})\leq \frac{1}{K}
    \end{cases}\label{eqt: performance}
    \\[0.9em]
    C=&\sum_{x^{\prime}\in \mathcal{X}}R+\lambda \mathrm{Sem}(x^{\prime},x),\label{eqt: normalizing constant}
\end{align}
where $K$ is the number of classes. In Eq. \ref{eqt: target dis}, we construct adversarial distribution by utilizing a hyper-parameter $\lambda$ to combine the attack performance $R$ and semantic similarity $Sem(\cdot)$. In Eq. \ref{eqt: normalizing constant}, $C$ represents the constant normalizing $\pi(x^{\prime})$ to ensure the distribution condition, $\sum_{x^{\prime}\in \mathcal{X}} \pi(x^{\prime}\vert x, \lambda)=1$. To keep more semantics, we let $\mathrm{Sem}(x^{\prime},x) $ denote the semantic similarity between adversarial example $x^{\prime}$ and original text $x$. In general, the $Sem(\cdot)$ is implemented with the cosine similarity between sentence encodings from a pre-trained sentence encoder, such as USE \cite{cer2018universal}.

\begin{table}\centering
\small
\caption{Datasets and accuracy of victim models before attacks.}
\begin{tabular}{cccccc}
\toprule
Dataset &Size  & Task      &  Model         &  Accuracy  \\
\midrule
\multirow{2}[0]{*}{AG's News}    &\multirow{2}[0]{*}{127000}
  &\multirow{2}[0]{*}{\shortstack{News\\ topics}}&
BERT-C &  94\%    \\
~&~&~     &  TextCNN   &  90\%  \\
\midrule
\multirow{2}[0]{*}{Emotion}  &
\multirow{2}[0]{*}{20000}    &\multirow{2}[0]{*}{\shortstack{Sentiment\\analysis}}&  BERT-C      &   97\%
\\
~  &~&~ & TextCNN &   93\%  \\
\midrule
\multirow{2}[0]{*}{SST2} & 
\multirow{2}[0]{*}{9613}   &\multirow{2}[0]{*}{\shortstack{Sentiment\\analysis}}
&BERT-C      &   91\%   \\
~&~&~ &  TextCNN   &  83\% \\

\midrule
\multirow{2}[0]{*}{IMDB} & 
\multirow{2}[0]{*}{50000}   &\multirow{2}[0]{*}{\shortstack{Movie\\review}}
&BERT-C      &   93\%   \\
~&~&~ &  TextCNN   &  88\% \\

\bottomrule
\end{tabular}

\label{tab: datasets and models}
\end{table}

\begin{table*}[tbh]
\small
\centering
\caption{The successful attack rate (SAR) of attack algorithms. The higher values of SAR indicate better performance. For each row, the highest SAR is highlighted in bold, the second highest SAR is highlighted by underline, and the third highest SAR is denoted with italic.}
\begin{tabular}{p{1.4cm}p{1.2cm}p{1.1cm}p{1.1cm}p{1.1cm}p{1.1cm}p{1.1cm}p{1.1cm}p{1.1cm}p{1.1cm}p{1.1cm}}
\toprule
\multirow{2}[3]{*}{Dataset}&\multirow{2}[3]{*}{Model}&  \multicolumn{8}{c}{Attack Methods} \\
\cmidrule(lr){3-11}
~& ~                     & WMP & A2T &BAE &FAGA  &BERT.A & CLARE & PWWS & PSO &FBA \\
\midrule
%%%%%% AG's News %%%%%%
\multirow{2}[0]{*}{AG's News}&BERT-C& 21.39\%&
15.25\%& 19.50\%&25.17\%  &47.71\% & \textit{71.32\%} &\underline{76.75\%} &67.76\%  &\textbf{81.90\%}\\

~ &TextCNN&25.19\%&
33.31\%& 39.13\% &56.10\% & 70.21\% &\textit{79.31\%} &\underline{85.31\%} &76.24\% &\textbf{93.12\%}  \\
\midrule
%%%%%% Emotion %%%%%%
\multirow{2}[0]{*}{Emotion}&BERT-C& 44.91\%&
48.06\% & 62.68\% &78.21\%  & 85.90\% &\underline{99.06\%} &91.75\% & \textit{94.76\%}&\textbf{99.15\%} \\

~ &TextCNN& 83.11\%&
81.11\%&85.34\%& 90.10\% & 95.11\%&97.11\% & \underline{98.20\%} &\underline{99.00\%} &\textbf{100\% }  \\
\midrule
%%%%%% SST2 %%%%%%
\multirow{2}[0]{*}{SST2}&BERT-C&52.43\%&
41.74\%& 55.14\% &77.07\% &77.21\% &\textit{94.79\%} &93.9\% &\underline{96.62\%} &\textbf{99.31\%}\\

~ &TextCNN&71.69\%&
77.41\%&80.15\% & 92.14\%& 83.23\% & 85.55\% &\underline{98.13\%} & \textit{92.20\%} &\textbf{100\%} \\
\midrule
%%%%%% IMDB %%%%%%
\multirow{2}[0]{*}{IMDB}&BERT-C& 
79.1\%&76.1\%&80.3\%&90.4\%&89.1\% & \textit{95.1\%}&94.1\%&\textbf{100.0\%}&\textbf{100.0\%}\\

~ &TextCNN&
77.0\% & 81.0\% & 89.3\% & 92.5\% & 91.1\% & 98.4\% & \textit{99.6\%} & \textbf{100.0\%}&\textbf{100.0\%}\\

\bottomrule
\end{tabular}

\label{tab: SAR}
\end{table*}
In spite of the use of the semantic regularizer, we argue that a high  $R$ might still cause a thrilling semantic loss because the value of $\pi(x^{\prime} \vert x)$ might go up with large increases of $R$ and small drops of semantic similarity $Sem(x^{\prime},x)$. Thus, for a further improvement on semantic preservation, we let the $R$ be associated with a cut-off value at $\frac{1}{K}$ when the class is successfully misclassified (i.e.,  when $F_{y}(x^{\prime})\leq \frac{1}{K}$). Note that when $F_{y}(x^{\prime})\leq \frac{1}{K}$, the classifier will misclassify  $x'$ to one of the other $K-1$ classess other than $y$. By having the mechanism of setting $R$ to $\frac{1}{K}$ whenever misclassification is achieved, all successful adversarial examples will have the same $R$ value. This way, their optimization with $\pi$ will then focus on maximizing their semantic similarity $Sem(x^{\prime},x)$ with the original texts.

\subsection{Fraud's Bargain Attack via Metropolis Hasting Sampler}
Metropolis Hasting simulates a target distribution by using a proposing function to offer a trial state which is then accepted or rejected according to a customized acceptance probability. Specifically, given a target distribution $Q(\cdot)$, the MH sampler utilizes a proposing function $q(s_{t+1}\vert s_{t})$ (transition density from $s_t$ to $s_{t+1}$) to construct a Markov Chain, whose equilibrium distribution is our target distribution. By this probabilistic mechanism, the proposing function would propose a trial state $s_{t+1}$ given the current state $s_t$ and the acceptance probability $\alpha(s_{t+1}\vert s_{t})$, as shown in Eq.  \ref{eqt: general accept prob}:
\begin{align}
    \alpha(s_i,s_{i+1})=\min \left(1, \frac{Q\left(s_{i+1}\right)}{Q(s_i)} \frac{q\left(s_i \mid s_{i+1}\right)}{q\left(s_{i+1} \mid s_i \right)}\right) \label{eqt: general accept prob}
\end{align}

Based on such a setting, we construct FBA by considering the adversarial distribution (Eq. \ref{eqt: target dis}) and the WMP as the MH's target distribution and proposing function, respectively. In each iteration of FBA, we use WMP to propose a trial state $x_{t+1}$ and calculate the acceptance probability $\alpha(x_{t+1}\vert x_{t})$. FBA's acceptance probability in Eq. \ref{eqt: general accept prob} can then be mathematically formulated by using WMP as follows:
\begin{align}
    \alpha(x_{t+1}\vert x_{t})=\min \left(1, \frac{\pi\left(x_{t+1}\right)}{\pi(x_t)} \frac{\mathrm{WMP}\left(x_t \mid x_{t+1}\right)}{\mathrm{WMP}\left(x_{t+1} \mid x_t, \right)}\right) \label{eqt: accept prob}
\end{align}
where $\mathrm{WMP}\left(x_t \mid x_{t+1}\right)$ is guaranteed above zero by Theorem \ref{theo: WMP reverse}, and calculated by reversing the WMP process: removing the inserted word, inserting the removed word and recovering substituted word. Once the value of $\alpha(x_{t+1}\vert x_{t})$ is computed, we proceed to sample $u$ from a uniform distribution ranging from 0 to 1, i.e., $u\sim Unif(0,1)$. If $u<\alpha(x_{t+1}\vert x_{t})$ we will accept $x_{t+1}$ as the new state, otherwise the state will remain as $x_t$. After performing $T$ iterations, FBA produces a collection of adversarial candidates, among which we select the one with the smallest modification that successfully flips the predicted class. The complete process of FBA is presented in Algorithm \ref{algo: FBA}.
% With such an accept probability, we expect the algorithm to asymptotically converge to the candidate which degrade the model performance while keeping enough semantics. 
With this acceptance probability, we expect the algorithm to asymptotically converge to a candidate that compromises the model's performance while preserving sufficient semantic integrity.

\begin{table*}[tbh]
\small
\caption{The imperceptibility performance (ROUGE, USE, Mod) of attack algorithms. The higher values of ROUGE and USE indicate better performance while the lower Mod indicates better performance. For each row, the best performance is highlighted in bold, the second best is highlighted in underline, and the third best is denoted with italic.}
\centering
\begin{tabular}{p{1.2cm}p{1.2cm}p{1.4cm}p{1cm}p{1cm}p{1cm}p{1cm}p{1cm}p{1cm}p{1cm}p{1cm}p{1cm}}
\toprule
\multirow{2}[3]{*}{Dataset}&\multirow{2}[3]{*}{Model} & \multirow{2}[3]{*}{Metrics} &  \multicolumn{7}{c}{Attack Methods} \\
\cmidrule(lr){4-12} 
~& ~                 & ~           & WMP     & A2T &BAE &FAGA & BERT.A & CLARE & PWWS & PSO &FBA \\
\midrule
\multirow{6}[0]{*}{\shortstack{AG's\\News}}&\multirow{3}[0]{*}{BERT-C}  & ROUGE $\uparrow$    & 0.5149
&0.8311&0.8145 &0.7721 &0.8003& 0.6673 &\textit{0.8356} &\textbf{0.8521}&\underline{0.8453}\\

~ &~                        & USE $\uparrow$     &0.7083
&0.7143 & 0.7234 &\textit{0.7534} &0.7334 & 0.6603 &0.7332 &\underline{0.7732}&\textbf{0.8001}\\

~ &~                        & Mod $\downarrow$  &17.2\%   & \textit{13.3\%} & \underline{11.3\%} & 14.6\% & 17.4\% & 14.2\% &17.9\% &21.6\% & \textbf{11.0\%} \\
\cmidrule{2-12}
~&\multirow{3}*{TextCNN}      & ROUGE $\uparrow$   &0.5149
&0.8291&0.8355 &0.7663 &\underline{0.8643}& 0.6969  &0.8366 &\textit{0.8401} &\textbf{0.8711}\\
~&~                         & USE $\uparrow$   &0.7083
&0.7542 & 0.7257 &0.8014 &\textit{0.8103} & 0.7121 &\underline{0.8111} &\textit{0.8103}&\textbf{0.8224}\\
~ &~                        & Mod $\downarrow$    & 19.2\% & 11.9\% & \textbf{10.3\%}  &\textit{11.5\%} & 15.4\% & 14.1\% & 16.5\% & 15.5\%  &\textbf{10.3\%} \\

\midrule

%%%%%%%%%%% Dataset: Emotion  %%%%%%%%%%%

\multirow{6}[0]{*}{Emotion}&\multirow{3}[0]{*}{BERT-C}  & ROUGE $\uparrow$     & 0.5021
&0.6111 & 0.5991 &0.5911 &\underline{0.6401} &0.4441 & 0.6141 &  \textit{0.6256} &  \textbf{0.6410 }\\

~ &~                        & USE $\uparrow$    &0.8803
&0.8951& 0.8831& 0.8940 & 0.8714 &0.8201 &\textit{0.9012} &\underline{0.9210} &\textbf{0.9246} \\

~ &~                        & Mod $\downarrow$     & 9.9\%
& \textit{9.1\%} & \underline{7.7\%}  &9.8\% & 9.2\% & 11.2\% &10.2\% & 12.0\%&\textbf{7.3\%}\\
\cmidrule{2-12}

~&\multirow{3}*{TextCNN}    & ROUGE $\uparrow$  &0.6210
&0.6891 & 0.6413 &\textit{0.6932} & 0.6201 &0.5623 & \underline{0.6994} &  0.6613 &  \textbf{0.7304}\\
~&~                         & USE $\uparrow$  & 0.7982
&\textit{0.8530}& 0.7341& 0.80412 & 0.8315 &0.5631 &\underline{0.8920} &0.8342 &\textbf{0.9046}\\
~ &~                        & Mod $\downarrow$    &13.0\% & 10.3\% & 9.8\% & \underline{8.3\%} & \textit{9.3\%} & 15.3\% & 19.0\% & 18.0\% & \textbf{8.0\%}\\

\midrule

%%%%%%%%%%% Dataset: SST2  %%%%%%%%%%%

\multirow{6}[0]{*}{SST2}&\multirow{3}[0]{*}{BERT-C}  & ROUGE $\uparrow$     & 0.6621
&0.6642 & \underline{0.7521} & \textit{0.7201} &0.6352 &0.5501 & 0.6341 &  0.7041 &  \textbf{0.7540}  \\
~ &~                        & USE $\uparrow$      & 0.7938
&0.8421& 0.7304& 0.7310 & 0.8564 &0.7434 &0.8422 &0.8104 &\textbf{0.8820}  \\
~ &~                        & Mod $\downarrow$   &18.1\%  &\textit{13.4\%} & \underline{11.3\%} & 17.2\% & 13.5\%  & 22.1\%  & 17.2\%& 17.2\%& \textbf{10.1\%} \\
\cmidrule{2-12}
~&\multirow{3}*{TextCNN}      & ROUGE $\uparrow$   &  0.6393
&0.6499 & \underline{0.7139} &0.6821 &0.6532 &0.5701 & 0.6393 &  \textit{0.7001} &  \textbf{0.7340}\\
~&~                         & USE $\uparrow$   & 0.6901
&0.7521& 0.7024& 0.6931 & 0.7410 &0.7009 &\underline{0.8210} &\textit{0.8162} &\textbf{0.8709}\\

~ &~                        & Mod $\downarrow$     &15.9\%
& 16.1\% & \underline{10.4\%} &15.3\% & \textit{13.4\%} &  17.1 \% & \textit{13.4\%} &17.2\% &\textbf{10.0\%} \\

\midrule

%%%%%%%%%%% Dataset: IMDB  %%%%%%%%%%%

\multirow{6}[0]{*}{IMDB}&\multirow{3}[0]{*}{BERT-C}  & ROUGE $\uparrow$     
&0.7934 &0.8123 & 0.8153 & \textit{0.8234} & 0.8112 &0.8233&\underline{0.8531}& 0.8021&\textbf{0.8621}\\
~ &~                        & USE $\uparrow$     
&0.8118&0.8561&0.8321&0.8246&\textit{0.8691}& 0.8369&\underline{0.9080}&0.8600&\textbf{0.9102}\\
~ &~                        & Mod $\downarrow$     &8.0\%&6.7\%&7.7\%&6.7\%&\textbf{5.9\%}&8.3\%&\textit{6.2\%}&9.2\%&\textbf{5.9\%}\\
\cmidrule{2-12}
~&\multirow{3}*{TextCNN}      & ROUGE $\uparrow$   
&0.8104 &0.8321 & 0.8611 & \textit{0.8814} & 0.8412 &0.8600&\underline{0.8821}& 0.8123&\textbf{0.9121}\\
~&~                         & USE $\uparrow$   
&0.8211& 0.8905 &0.8613 & 0.7714 & \textit{0.8911}& 0.7979 &\underline{0.9000}&0.8600&\textbf{0.9092}\\

~ &~                        & Mod $\downarrow$     
&7.1\% & 7.1\% & 8.7\% & \underline{4.7\%} & 5.5\% & 6.3\% & \textit{5.2\%} & 7.6\% &\textbf{3.9\%}\\

\bottomrule
\end{tabular}

\label{tab: imper}
\end{table*}

\begin{table*}[tbh]
\small
\caption{The quality of generated adversarial examples measured by fluency (PPL) and grammar error rate (GErr) of attack algorithms. The lower values of PPL and GErr indicate better performance. For each row, the best quality is highlighted in bold, the second best quality is highlighted by underline, and the third best quality is denoted with italic.}
\centering
\begin{tabular}{p{1.1cm}p{1.2cm}p{1.2cm}p{1cm}p{1cm}p{1cm}p{1cm}p{1cm}p{1cm}p{1cm}p{1cm}p{1cm}}
\toprule
\multirow{2}[3]{*}{Dataset}&\multirow{2}[3]{*}{Model} & \multirow{2}[3]{*}{Metrics} &  \multicolumn{8}{c}{Attack Methods} \\
\cmidrule(lr){4-12} 
~& ~                 & ~                  &WMP & A2T &BAE &FAGA & BERT.A & CLARE & PWWS & PSO &FBA \\
\midrule
\multirow{4}[0]{*}{\shortstack{AG's\\News}}&\multirow{2}[0]{*}{BERT-C}  & PPL $\downarrow$   &331
&291 & \underline{142} & \textit{165} & 281 & 233 &321 & 292 & \textbf{133}\\

~ &~                        & GErr $\downarrow$   & 0.22
&\underline{0.19}  & \underline{0.19} & \underline{0.19} & \underline{0.19} &0.21 & 0.22& 0.31 & \textbf{0.18}\\
\cmidrule{2-12}
~&\multirow{2}*{TextCNN}      & PPL $\downarrow$   & 281
&199 & \underline{142} &145 &241 & 213 &277  & \textit{143} & \textbf{133}\\
~&~                         & GErr $\downarrow$   &0.23
&0.17 & 0.19 & \textit{0.15} & \textit{0.15} & 0.19 & 0.20 & \underline{0.14} &\textbf{0.13}\\
\midrule

%%%%%%%%%%% Dataset: Emotion  %%%%%%%%%%%

\multirow{4}[0]{*}{Emotion}&\multirow{2}[0]{*}{BERT-C}  & PPL $\downarrow$     & 281
&\textit{251} & \underline{233} & 259 &  301& 298  & 333 &  336 &  \textbf{229}\\

~ &~                        & GErr $\downarrow$    &0.19
&0.19& \textbf{0.10} & \textit{0.13} & 0.17 & 0.14 &0.16 & 0.20 &\textbf{0.10} \\

\cmidrule{2-12}
~&\multirow{2}*{TextCNN}    & PPL $\downarrow$   & 288
&268 & \underline{201} & 241 & \textit{221} & 256 & 299&  301&  \textbf{182}\\
~&~                         & GErr $\downarrow$   & 0.16
&\textit{0.11}& \textbf{0.10}& 0.13 & 0.13 & 0.16 &0.19 &0.18 & \textbf{0.10}\\
\midrule

%%%%%%%%%%% Dataset: SST2  %%%%%%%%%%%

\multirow{4}[0]{*}{SST2}&\multirow{2}[0]{*}{BERT-C}  & PPL $\downarrow$     &213
&211 & \textit{173} & 182 & 200 &\underline{155} & 214 &  197 &  \textbf{142}  \\
~ &~                        & GErr $\downarrow$     &0.20
&\textit{0.18}& \underline{0.15} & 0.24 & 0.25 & 0.19 & 0.21 &0.27 &\textbf{0.14}  \\
\cmidrule{2-12}
~&\multirow{2}*{TextCNN}      & PPL $\downarrow$   &198
&164 & \underline{142} &163 & 210 & 174 &224 & \textit{145} & \textbf{140}\\
~&~                         & GErr $\downarrow$  &0.19
&0.21& \textit{0.14} &  0.17 & 0.21 &0.17 &0.23 &\textbf{0.13} &\textbf{0.13}\\

\midrule

%%%%%%%%%%% Dataset: IMDB  %%%%%%%%%%%

\multirow{4}[0]{*}{IMDB}&\multirow{2}[0]{*}{BERT-C}  & PPL $\downarrow$     
&91&\underline{62}&\textit{70}&90&83&101&88&90&\textbf{60}  \\
~ &~                        & GErr $\downarrow$     
&0.28&0.22&\textit{0.19}&0.21&\textit{0.19}&\textbf{0.18}&\textit{0.19}&0.22&\textbf{0.18}\\
\cmidrule{2-12}
~&\multirow{2}*{TextCNN}      & PPL $\downarrow$   
&103&\underline{69}&79&99&80&89&\underline{69}&99&\textbf{66}\\
~&~                         & GErr $\downarrow$  
&0.33&0.29&0.22&0.26&0.23&\textbf{0.20}&\textbf{0.20}&0.28&\textbf{0.20}\\

\bottomrule
\end{tabular}

\label{tab: quality}
\end{table*}

\section{Experiments and Analysis}
\label{experiments}

\noindent We assess the performance of our methods using popular and publicly accessible datasets, along with high-performing victim classifiers. In order to facilitate reproducibility, we have made our code and data available on a GitHub repository \footnote{ \url{https://github.com/MingzeLucasNi/FBA}.}

\subsection{Datasets and Victim Models}
% In this subsection, we detail the three benchmark datasets, and the two well-performed textual classifiers.
\subsubsection{Dataset} We conduct experiments on four publicly accessible benchmark datasets. AG’s News \cite{ag_news} is a news classification dataset with 127,600 samples belonging to 4 topic classes. Emotion \cite{emotion} is a dataset with 20,000 samples and 6 classes. SST2 \cite{sst2} is a binary class topic dataset with 9,613 samples. IMDB \cite{sst2} is a binary class topic dataset with 50,000 labeled samples. Details of these datasets can be found in Table \ref{tab: datasets and models}.
\par
\subsubsection{Victim Models} We apply our attacks to two popular and well-performed victim models below.
\paragraph*{BERT-based Classifiers} 

To perform experiments, three BERT-based models, collectively called BERT-C models, were selected and pre-trained by Huggingface. The structures of the classifiers were adjusted based on the dataset sizes. BERT classifiers were trained on AG's News and Emotion datasets using Distil-RoBERTa-base and Distil-BERT-base-uncased, respectively. The smaller version of BERT, BERT-base-uncased, was used for the SST2 dataset. The accuracy of the models before the attack can be found in Table \ref{tab: datasets and models}, and the models are publicly available for reproducibility purposes.

\paragraph*{TextCNN-based models}
The TextCNN victim model is a different type of model from the BERT-based models. It has a 100-dimension embedding layer followed by a 128-unit long short-term memory layer. This classifier is trained 10 epochs by ADAM optimizer with the default hyper-parameter in Pytorch.The accuracy of these TextCNN-base models is also shown in Table \ref{tab: datasets and models}.

% \begin{table*}[t]
%     \centering
%     \caption{Comparisons between the FBA and its ablation WMP on AG's News dataset. The better performance is highlighted in bold.}
%     \begin{tabular}{p{0.9cm}p{0.95cm}p{1cm}p{0.95cm}p{0.95cm}p{0.85cm}p{0.85cm}p{1cm}p{1cm}p{0.9cm}p{0.9cm}p{0.85cm}p{0.85cm}}
%     \toprule
%          \multirow{2}[3]{*}{Method}&\multicolumn{6}{c}{BERT-C}&\multicolumn{6}{c}{TextCNN}\\ \cmidrule(lr){2-7} \cmidrule(lr){8-13}
%          ~&SAR$\uparrow$&ROUGE$\uparrow$&USE$\uparrow$&Mod$\downarrow$&PPL$\downarrow$&GErr$\downarrow$&   SAR $\uparrow$&ROUGE$\uparrow$&USE$\uparrow$&Mod$\downarrow$&PPL$\downarrow$&GErr$\downarrow$  \\ \midrule
%          WMP&21.39\% &0.5149&0.7083&17.2\% & 331& 0.22&
%          25.19\%&0.5149&0.7083& 19.2\% & 281 &0.23 \\

%          FBA&\textbf{81.90\%}& \textbf{0.8453}& \textbf{0.8001}& \textbf{11.0\%} & \textbf{133} & \textbf{0.18}&
%          \textbf{93.12\%}& \textbf{0.8711} &\textbf{0.8224} & \textbf{10.3\%}& \textbf{133} & \textbf{0.13}\\
%     \bottomrule
%     \end{tabular}
%     \label{tab: ablation}
% \end{table*}

\begin{table*}[th]
\small
\caption{Adversarial examples of Emotion dataset for victim classifier BERT-C. Blue texts are original words, while red ones are substitutions. The true class scores is placed inside the brackets. The lower true class score indicates better performance. The best attacks is bold.\cite{mingze2023}}
\centering
\begin{tabular}{m{3cm}|m{13cm}}
\toprule
Attacks & Adversarial examples \\  \midrule
A2T (Unsuccessful attack. True class score = 41.31\%)& 
i spent wandering around still kinda dazed and not \textcolor{blue}{\st{feeling}} \textcolor{red}{sense}  particularly \textcolor{blue}{\st{sociable}} \textcolor{red}{social} but because id been in hiding for a couple for days and it was getting to be a little unhealthy i made myself go down to the cross and hang out with folks
\\ \midrule
BAE (Unsuccessful attack. True class score = 33.25\%)& 
i spent wandering around still kinda dazed and not \textcolor{blue}{\st{feeling}} \textcolor{red}{being} particularly sociable but because id been in hiding for a couple for days and it was getting to be a \textcolor{blue}{\st{little}} \textcolor{red}{bit} unhealthy i made myself go down to the cross and hang out with folks
\\ \midrule
FAGA (\textbf{Successful attack}. True class score = 13.32\%)& 
i spent wandering around still kinda dazed and not feeling particularly \textcolor{blue}{\st{sociable}} \textcolor{red}{sympathetic} but because id been in hiding for a \textcolor{blue}{\st{couple}} \textcolor{red}{few} for days and it was getting to be a little unhealthy i made myself go down to the cross and hang out with folks
\\ \midrule
BERT.A (Unsuccessful attack. True class score = 77.04\%)& i spent wandering around still kinda dazed and not \textcolor{blue}{\st{feeling}} \textcolor{red}{being} particularly \textcolor{blue}{\st{sociable}} \textcolor{red}{happy} but because id been in hiding for a couple for days and it was getting to be a little unhealthy i made myself go down to the cross and hang out with folks\\ \midrule
CLARE (\textbf{Successful attack}. True class score = 10.54\%)& 
i spent wandering around still kinda dazed and not feeling particularly \textcolor{blue}{\st{sociable}} \textcolor{red}{lonely} but because id been in hiding for a couple for days and it was getting to be a little unhealthy i made myself go down to the cross and hang out with folks.
\\ \midrule
PWWS (\textbf{Successful attack}. True class score = 21.11\%)& 
i spent wandering around still kinda dazed and not \textcolor{blue}{\st{feeling}} \textcolor{red}{palpate} particularly sociable but because id been in hiding for a couple for days and it was getting to be a little unhealthy i made myself go down to the cross and hang out with folks
\\ \midrule
PSO (\textbf{Successful attack}. True class score = 6.90\%) & 
i spent wandering around still kinda dazed and not \textcolor{blue}{\st{feeling}} \textcolor{red}{considering} particularly sociable but because id been in hiding for a couple for days and it was getting to be a little unhealthy i made myself go down to the cross and hang out with \textcolor{blue}{\st{folks}} \textcolor{red}{dudes}
\\ \midrule
FBA (\textbf{Successful attack}. True class score = \textbf{0.63\%}) & i spent wandering around still kinda dazed and not \textcolor{blue}{\st{feeling}} \textcolor{red}{sensing} particularly sociable but because id been in hiding for a couple for days and it was getting to be a little unhealthy i made myself go down to the cross and hang out with folks\\ 
\bottomrule
\end{tabular}

\label{tab: example}
\end{table*}

\begin{table*}[hbt!]
\small
\caption{Adversarial examples of SST2 dataset for victim classifier TextCNN. Blue texts are original words, while red ones are substitutions. The true class scores is placed inside the brackets. The lower true class score indicates better performance. The best attack is bold.}
\centering
\begin{tabular}{m{4.6cm}|m{13cm}}
\toprule
Attacks & Adversarial examples \\  \midrule
A2T (\textbf{Successful attack}. True class score = 21.31\%)& 
an often-deadly \del{boring} \nt{short}, strange reading of a classic whose witty dialogue is \del{treated} \nt{laced} with a baffling casual approach
\\ \midrule
BAE (Unsuccessful attack. True class score = 32.10\%)& 
an often-deadly boring , strange reading of a \del{classic} \nt{novel} whose witty \del{dialogue} \nt{commentary} is treated with a baffling casual approach
\\ \midrule
FAGA (Unsuccessful attack. True class score = 51.12\%)& 
an \del{often-deadly} \nt{extremely} boring , strange reading of a \del{classic} \nt{characters} whose witty dialogue is treated with a baffling casual approach
\\ \midrule
BERT.A (Unsuccessful attack. True class score = 63.41\%)& an often-deadly boring , \del{strange} \nt{critical} reading of a classic whose \del{witty} \nt{spoken} dialogue is treated with a baffling casual approach
\\ \midrule
CLARE (Unsuccessful attack. True class score = 10.54\%)& 
an often-deadly boring , strange reading of a classic whose \del{witty} \nt{entire} dialogue is treated with a baffling \del{casual approach} \nt{casualness}
\\ \midrule
PWWS (Unsuccessful attack. True class score = 51.11\%)& 
an often-deadly boring, \del{strange} \nt{casual} reading of a classic whose witty dialogue is treated with a \del{baffling} \nt{more} casual approach
\\ \midrule
PSO (\textbf{Successful attack}. True class score = 7.90\%) & 
an often-\del{deadly} \nt{somewhat} boring, \del{strange} \nt{humorous} reading of a classic whose witty dialogue is treated with a baffling casual approach
\\ \midrule
FBA (\textbf{Successful attack}. True class score = \textbf{5.11\%}) & an \del{often-deadly} \nt{often-harmful} boring, strange reading of a classic whose witty dialogue is treated with a baffling casual approach\\ 
\bottomrule
\end{tabular}

\label{tab: example3}
\end{table*}

\subsection{Baselines}
We choose the following solid baselines:
\begin{itemize}
    \item Faster Alzantot Genetic Algorithm (FAGA) \cite{jia2019faga} accelerate Alzantot Genetic Algorithm \cite{alzantot2018generating}, by bounding the searching domain of genetic optimization.
    \item BAE \cite{garg2020bae} replaces and inserts tokens in the original text by masking a portion of the text and leveraging the BERT-MLM.
%to generate alternatives for the masked tokens
    \item BERT-Attack \cite{li2020bertattack} takes advantage of BERT MLM to generate candidates and attack words by the static WIR descending order.
    \item A2T \cite{yoo2021a2t} uses a  gradient-based word importance ranking method to iteratively replace each word with synonyms generated from a counter-fitted word embedding.% \cite{mrksic-etal-2016-counter}
    \item CLARE \cite{li2021clare}  applies a sequence of contextualized perturbation actions to the input. Each can be seen as a local mask-then-infill procedure: it first applies a mask to the input around a given position and then fills it in using a pre-trained MLM.
    \item PWWS \cite{ren2019pwws} method proposed in Ren et al. (2019) selects candidate words from WordNet, a large lexical database of English, and then sorts them in attack order by multiplying the word saliency with the probability variation.   
    \item Particle Swarm Optimization (PSO)\cite{zang2020word} is a method that selects word candidates from HowNet \cite{dong2003hownet} and uses PSO to find adversarial text via combinatorial optimization.

\end{itemize}
\subsection{Evaluation Metrics and Experimental Setting}
We use the following six metrics to measure the performance of adversarial attacks.
\begin{itemize}
    \item Successful attack rate (SAR) denotes the percentage of adversarial examples that can successfully attack the victim model.
    \item Universal Sentence Encoder (USE) \cite{cer2018universal} measures the semantic similarity by calculating the cosine similarity between the input and its adversary.
    \item Modification Rate(Mod) is the percentage of modified tokens. Each replacement, insertion or removal action accounts for one modified token.
    \item  ROUGE: measures the overlap of n-grams between the candidate and reference sentences, to evaluate the alignment similarity between adversarial examples and original sentences.
    \item Grammar Error Rate (GErr) is measured by the absolute rate of increased grammatic errors in the successful adversarial examples, compared to the original text, where we use LanguageTool \cite{naber2003rule} to obtain the number of grammatical errors.
    \item Perplexity (PPL) denotes a metric used to evaluate the fluency of adversarial examples and is broadly applied in the literature \cite{li2021clare,zang2020word}.
\end{itemize}
The SAR is tailored for assessing attack performance, focusing on the imperceptibility of adversarial examples via semantic similarity, modification extent, and word order. We utilize USE for semantic analysis, Mod to quantify modifications, and ROUGE to evaluate syntactical structure. Fluency is measured using GErr for grammatical accuracy and PPl for readability, ensuring the adversarial examples are seamless and indistinguishable.
\par
For settings of FBA, we set WMP action proposing probability in Eq. \ref{eqt: action} as $\mathbf{P}_{ins}=0.2$, $\mathbf{P}_{sub}=0.6$, and $\mathbf{P}_{rem}=0.2$. While setting up the distribution for selecting the substitution and insertion words, we believe MLM and Synonyms are equally important. Therefore we set the weights of these two methods  equal by setting $a_1=0.1,a_2=0.4,a_3=0.5$ and $b_1=0.5,b_2=0.5$ from Eq. \ref{eqt: sub dis} and Eq. \ref{eqt:insert}, respectively. In addtion, we only consider examples with a USE greater than 0.5 and where the modification rate remains below 0.25. We believe that an attack which modifies too many words or significantly alters the semantics, even if successful, cannot be deemed a quality attack

\subsection{Experimental Results and Analysis}
The experimental results of SAR and the imperceptibility performance (ROUGE, USE) and sentence quality (GErr, PPL) are listed in Table \ref{tab: SAR} and Table \ref{tab: imper} and Table \ref{tab: quality}, respectively. To give an intuitive of the generated examples, we also show two generated adversarial examples in Table \ref{tab: example} and \ref{tab: example3}. We manifest the three contributions mentioned in the Introduction section by asking four research questions:
\begin{table*}[t]
\small
\caption{Targeted attack results on Emotion dataset. The better-performed attack is highlighted in bold.}
\centering
\begin{tabular}{p{1.2cm}p{0.8cm}p{0.8cm}p{0.8cm}p{0.8cm}p{0.8cm}p{0.8cm}p{0.8cm}p{0.8cm}p{0.8cm}p{0.8cm}p{0.8cm}p{0.8cm}}
\toprule
\multirow{2}[3]{*}{Model} & \multicolumn{2}{c}{SAR $\uparrow$} &  \multicolumn{2}{c}{ROUGE $\uparrow$} &\multicolumn{2}{c}{USE $\uparrow$} &\multicolumn{2}{c}{Mod $\downarrow$} &\multicolumn{2}{c}{PPL $\downarrow$} &\multicolumn{2}{c}{GErr $\downarrow$} \\
\cmidrule(lr){2-3} \cmidrule(lr){4-5} \cmidrule(lr){6-7} \cmidrule(lr){8-9} \cmidrule(lr){10-11} \cmidrule(lr){12-13}
~           & PWWS        & FBA              & PWWS       & FBA & PWWS        & FBA  & PWWS        & FBA & PWWS        & FBA & PWWS        & FBA      \\
\midrule
BERT-C    & $21.23\%$   & \textbf{57.21\%}     & 0.4541     & \textbf{0.5101}&0.6012&\textbf{0.7732} & 13.1\% & \textbf{11.3\%} & 341  & \textbf{299}  &  0.28 & \textbf{0.22}\\
TextCNN    & $32.61\%$   & \textbf{65.07\%}    & 0.5603    & \textbf{0.6198}&0.6320&\textbf{0.6511} & 21.3\% & \textbf{15.1\%} & 411  & \textbf{223}  &  0.29 & \textbf{0.28}\\
\bottomrule
\end{tabular}

\label{tab: target}
\end{table*}

\begin{figure}[t]
    \centering
    \includegraphics[width=\columnwidth]{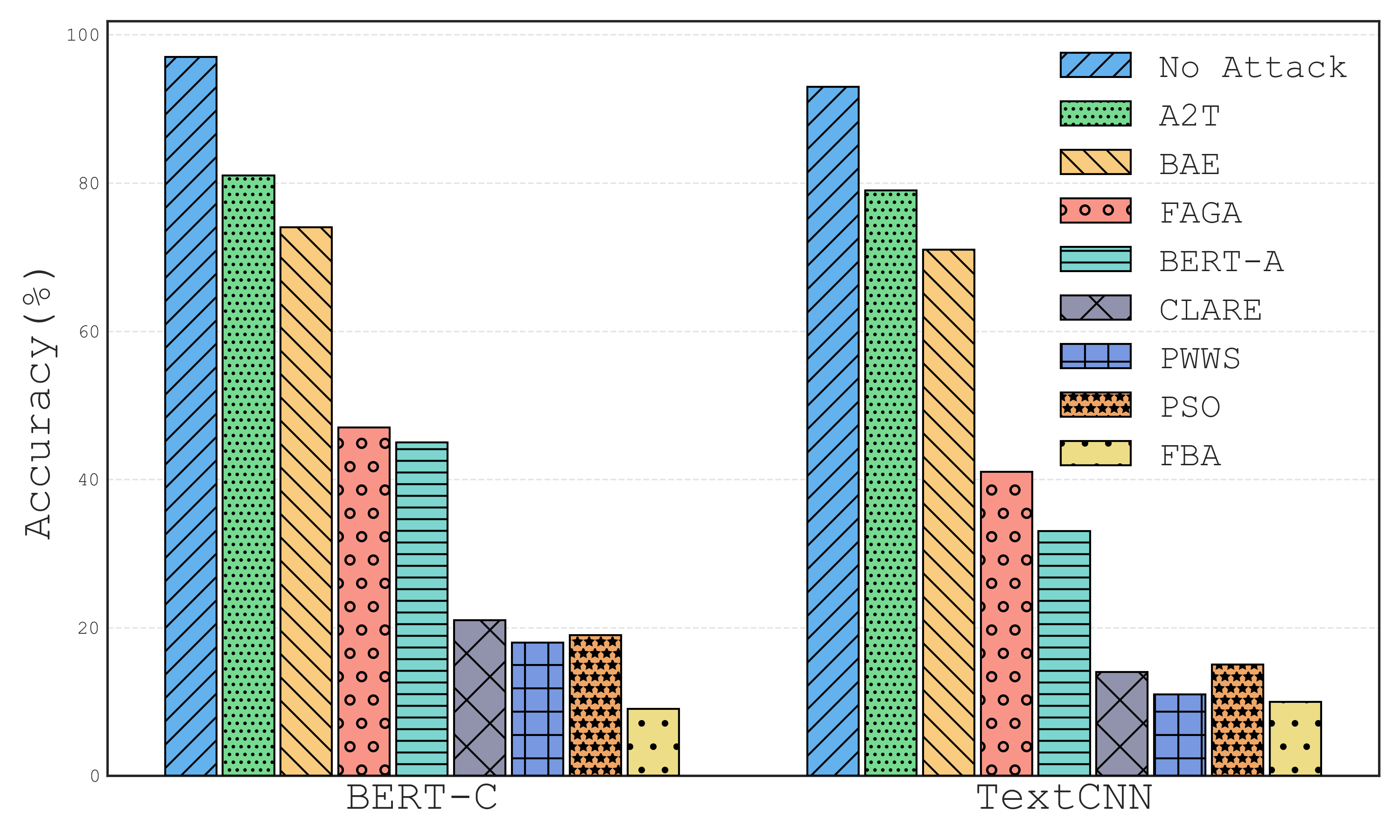}
    \caption{Performance of transfer attacks to victim models (BERT-C and TextCNN) on Emotion. Lower accuracy of the victim models indicates higher transfer ability (the lower the better).}
    \label{fig: transfer}
\end{figure}
\subsubsection*{(a) Does our FBA method make more thrilling attacks to baselines?}
We compare the attacking performance of the proposed  FBA method and baselines in Table \ref{tab: SAR}. To be more specific, Table \ref{tab: SAR} demonstrates that FBA consistently outperforms other competing methods across different data domains, regardless of the structure of classifiers. It can thus be concluded that the proposed method FBA achieves the best-attacking performance, with the largest successful attack rate (SAR). 
We attribute such an outstanding attacking performance to the two prevailing aspects of FBA. Firstly, the proposed FBA could enlarge the searching domain by removing, substituting and inserting words compared with the strategies with only substitution such that FBA provides more possible attacking combinations. Secondly, FBA optimizes the performance by stochastically searching the domain. Most of the baselines perform a deterministic searching algorithm with Word Importance Rank (WIR) could get stuck in the local optima. Differently, such a stochastic mechanism helps skip the local optima and further maximize the attacking performance.
\par    
\begin{figure}
    \centering
    \includegraphics[width=\columnwidth]{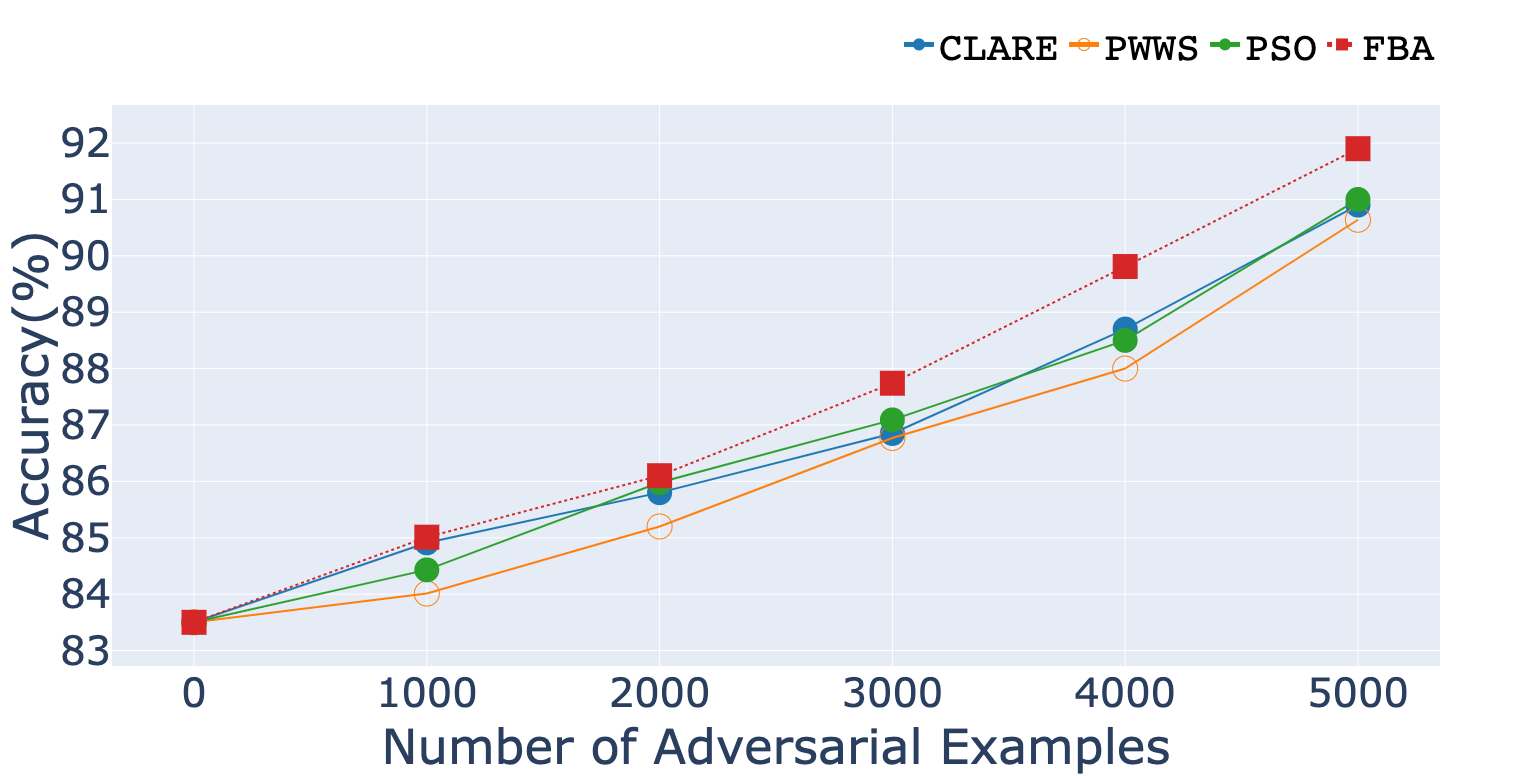}
    \caption{Retraining accuracy of TextCNN with different numbers of  adversarial examples included in the retraining. The higher the accuracy, the better the performance of the retraining.}
    \label{fig: adv_training}
\end{figure}
\begin{figure*}[tbh!]
    \centering
    \includegraphics[width=\textwidth]{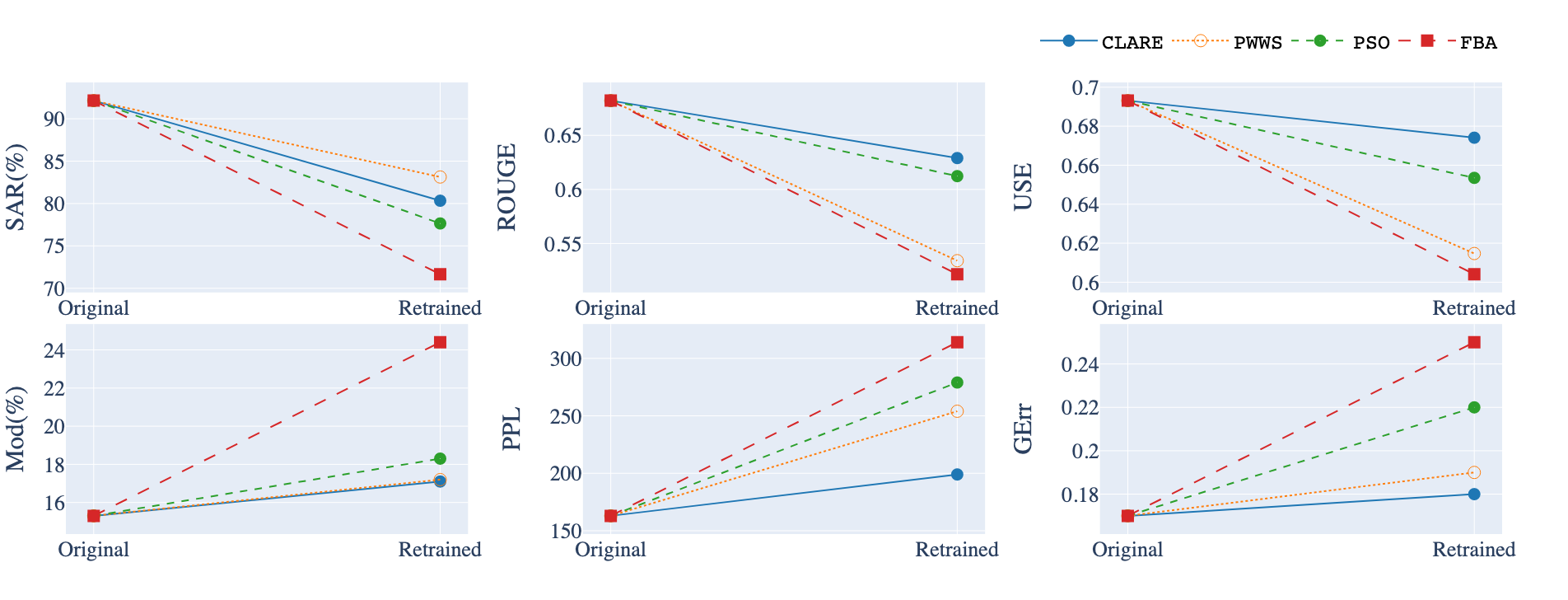}
    \caption{We employ FAGA to attack the adversarial retrained TextCNNs which joins adversarial examples from different attacking strategies (CLARE, PWWS, PSO and FBA) to the training set of SST2. The lower metrics (SAR, ROUGE, USE) suggest a better performance in robustness while The higher metrics (Mod, PPL, GErr) suggest a better performance in robustness.}
    \label{fig:retraining_robust}
\end{figure*}
\subsubsection*{(b) Is FBA superior to the baselines in terms of imperceptibility?}
We assess the imperceptibility of various attack strategies based on their semantic similarity (USE), modification rate (Mod), and word alignment (ROUGE) compared to the original text. Our findings, presented in Table \ref{tab: imper}, show that the FBA approach is generally more effective than the baseline methods. While attacking BERT-C on AG's News, FBA performs similarly to PSO in terms of ROUGE, but maintains a higher semantic similarity (USE) on that dataset, suggesting that PSO's performance comes at the expense of imperceptibility. Hence, FBA outperforms the other methods in terms of imperceptibility. This is mainly due to the customized target distribution, which helps to avoid excessive modifications, and the use of both MLM and kNN to find more semantically similar substitutions.

\par
\subsubsection*{(c) Is the quality of adversarial examples generated by the FBA 
better than that crafted by the baselines?}
High-quality adversarial examples should be parsing-fluent and grammarly correct. From Table \ref{tab: quality}, we can find that FBA provides the lowest perplexity (PPL), which  means the examples generated by FBA are more likely to appear in the corpus of evaluation. As our corpus is long enough and the evaluation model is broadly used, it indicates these examples are more likely to appear in natural language space, thus eventually leading to better fluency. For the grammar errors, the proposed method FBA is substantially better than the other baselines, which indicates better quality of the adversarial examples. We attribute such performance to our method of finding word substitution, constructing the candidates set by applying both MLM and kNN for synonym searching.
\par

% \subsubsection*{(d) Does the Metropolis-Hasting (MH) algorithm benefit the selection of the best adversarial candidates?}
% To test the performance of the Metroplis-Hasting algorithm, we did an ablation study by making a comparison between FBA and WMP whose adversarial candidates are not selected by the Metropolis-Hasting algorithm. Specifically, we perform these two attacks, FBA and WMP, to two classifiers, BERT-C and TextCNN pre-trained on dataset AG's News, and the experimental results are shown in Table \ref{tab: SAR}. From Table \ref{tab: ablation}, FBA achieved better performance in both attack (SAR), imperceptibility (USE, Mod, ROUGE) and sentence quality (PPL, GErr), thus we can conclude that the Metropolis-Hasting algorithm is effective in selecting the adversarial candidates.

\subsection{Ablation Studies}
\subsubsection*{(a) Evaluating the Effectiveness of MH}
To test the performance of the Metroplis-Hasting algorithm, we did an ablation study by making a comparison between FBA and WMP whose adversarial candidates are not selected by the Metropolis-Hasting algorithm. Specifically, we perform these two attacks, FBA and WMP, to two classifiers, BERT-C and TextCNN pre-trained on dataset AG's News, and the experimental results are shown in Table \ref{tab: SAR}, \ref{tab: imper} and \ref{tab: quality}. From these tables , FBA achieved better performance in both attack (SAR), imperceptibility (USE, Mod, ROUGE) and sentence quality (PPL, GErr), thus we can conclude that the Metropolis-Hasting algorithm is effective in selecting the adversarial candidates.

\subsubsection*{(b) Evaluating the Impact of Actions: Removal, Insertion, and Substitution}
To evaluate the influence of different actions on attack performance, we set up six distinct pairs with associated probabilities, as outlined in Table \ref{tab: ablation action}. In order to achieve equilibrium between the actions of removal and insertion, we consistently set their probabilities to be equal, represented as $\mathbf{P}_{ins} = \mathbf{P}_{rem}$.

The analysis of results from Table \ref{tab: ablation action} can be categorized into three key facets: attack performance (SAR), imperceptibility (ROUGE, USE, Mod), and the quality of the generated examples (PPL, GErr). For attack performance, the group with \( \mathbf{P}_{sub} = 0.4 \) exhibited the highest efficacy. Meanwhile, the groups with \( \mathbf{P}_{sub} = 0.6 \) and \( \mathbf{P}_{sub} = 0.8 \) trailed closely, differentiated by a slim margin. The initial three groups underperformed, primarily due to constraints that only consider examples with a USE exceeding 0.5 and a modification rate below 0.25. Hence, even though these groups could generate adversarial candidates capable of deceiving the target models, such examples are not regarded as successful adversarial instances. Furthermore, the group focusing solely on substitution also showcased a commendable success rate against overall performance.

Regarding imperceptibility, we observed an initial increase in performance with a rise in the substitution probability \( \mathbf{P}_{sub} \), which later began to decline. This trend can be attributed to the notion that a higher likelihood of insertions and removals can impact imperceptibility. Specifically, inserting or removing words may compromise language semantics, alignment, and parsing, as these actions can introduce significant losses or semantic redundancies. Simultaneously, if attackers focus exclusively on substitution, imperceptibility may suffer due to altering a larger set of words to bolster the attack's success. Hence, an optimal substitution probability likely exists that harmoniously balances imperceptibility. As for the quality of adversarial examples, there's a distinct pattern: the greater the substitution probability, the higher the quality. The experimental data suggests that inserting and removing affect sentence quality.

Our findings indicate that the performance is suboptimal when focusing solely on substitution or when excluding substitution altogether. This underscores the importance of considering all actions – substitution, removal, and insertion – to bolster the attack's effectiveness. It's imperative to gauge the overall success of adversarial attacks across three dimensions: attack potency, imperceptibility, and sentence quality. Engaging in adversarial attacks often necessitates trade-offs between imperceptibility and sentence quality, as documented in \cite{mingze2023,tradeoff}. Given varying attack objectives, attackers can adjust the substitution probability. Based on our experiments, a substitution probability of 0.6 (denoted as \( \mathbf{P}_{sub} = 0.6 \)) is recommended, as it strikes an ideal balance between attack efficacy and imperceptibility without undermining the textual quality.

\begin{table}
    \centering
    \caption{Performance comparison of FBA on the AG News dataset against the TextCNN victim model using different action probabilities. The top three performances are highlighted in bold, underlined, and italics.}
    \begin{tabular}{p{0.4cm}p{0.5cm}p{0.7cm}p{0.8cm}p{0.7cm}p{0.55cm}p{0.55cm}p{0.5cm}}
    \toprule
    $\mathbf{P}_{sub}$ & $\mathbf{P}_{ins}$& SAR & ROUGE & USE & Mod & PPL &GErr \\ \midrule
        0.0 & 0.5 & 69.12\% & 0.6711 & 0.7411 & 16.3\% & 231 & 0.21\\ 
        0.2 & 0.4 & 78.10\% & 0.7011 & 0.7524 & 15.3\% & 201 & 0.19\\ 
        0.4 & 0.3 &\textbf{94.12\%} & 0.7911 & 0.7771 & 15.1\% & 179 & 0.17\\ 
        0.6 & 0.2& \underline{93.12\%} & \textbf{0.8711} & \textit{0.8224} & \textbf{10.3\%} & \textit{133} & \underline{0.13}\\ 
        0.8 & 0.1 & \textit{91.12\%} & \underline{0.8211} & \textbf{0.8401} & \underline{12.3\%} & \underline{112} & \underline{0.13}\\ 
         1.0 & 0.0 & 87.01\% & \textit{0.8011} & \underline{0.8333} & \textit{14.3\%} & \textbf{110} & \textbf{0.12}\\ 
         \bottomrule
    \end{tabular}
    \label{tab: ablation action}
\end{table}

\subsubsection*{(c) Evaluating the Effectiveness of Word Candidates Selection}
Choosing the appropriate word candidates for substitution and insertion actions is crucial, as it directly impacts the success rate of attacks and imperceptibility. To evaluate the effectiveness of our word candidates selection method, we undertook an ablation study. This study compared performances utilizing a thesaurus (specifically, WordNet\cite{miller1990wordnet}), Masked Language Model (MLM), and Nearest Neighbors (NN) under L1, L2, and infinite norms. As shown from Table \ref{tab: ablation words}, our proposed method (MLM+$L_{1}$, MLM+$L_{2}$, MLM+$L_{\inf}$) for word candidate search exhibited superior performance than the baselines. We attribute this success to two main factors. Firstly, WMP utilizes NN to identify `potential' synonyms that, although not always precise, capture the desired meaning. Secondly, the MLM is crucial in improving the sentence's parsing structure. Furthermore, our observations indicate that the L2 norm marginally surpasses the L1 norm and significantly outperforms the infinite norm. While NN methods are not limited to any particular norm, our experimental results demonstrate the commendable efficacy of both L1 and L2 norms.

\begin{table}
    \centering
    \caption{Performance metrics for FBA against the TextCNN model on the AG News dataset using varied word candidate selection methods. The best three performances for each metric are highlighted in bold, underline, and italics.}
    \begin{tabular}{ccccccc}
    \toprule
         Methods & SAR & ROUGE & USE & Mod & PPL &GErr \\ \midrule
          WordNet & 55.12\% & 0.7103 & 0.7231 & 15.1\% & 260 & 0.21\\   
          $L_{\inf}$ & 63.06\% & 0.7401 & 0.7010 & 14.8\% & 281 & 0.20\\  
          $L_1$  & 70.19\% & 0.7419 & 0.7533 & 14.1\% & 209 & 0.18\\  
          $L_2$  & 72.88\% & 0.7812 & 0.7695 & 14.0\% & 201 & 0.19\\  
          MLM & 81.12\% & 0.7731 & 0.7031 & 15.3\% & 178 & \textit{0.17}\\ \midrule
          MLM+$L_{\inf}$ &\textit{88.12\%} & \textit{0.8441} & \textit{0.8001} & \textit{13.3\%} & \textit{140} & \textit{0.17}\\ 
          MLM+$L_1$ &\underline{92.42\%} & \textbf{0.8713} & \underline{0.8194} & \underline{11.1\%} & \underline{136} & \textbf{0.14}\\ 
          MLM+$L_2$ &\textbf{93.12\%} & \underline{0.8711} & \textbf{0.8224} & \textbf{10.3\%} & \textbf{133} & \underline{0.13}\\ 
          \bottomrule
    \end{tabular}
    \label{tab: ablation words}
\end{table}

\begin{table*}[t]
    \centering
    \caption{The time efficiency of attack algorithms evaluated with BERT-C on the Emotion and IMDB dataset. The metric of efficiency is second per example, which means a lower metric indicates a better efficiency. The horizontally best 3 methods will be bold, underlined and italic.}
    \begin{tabular}{cccccccccc}
    \toprule
        Datasets & WMP     & A2T &BAE &FAGA & BERT.A & CLARE & PWWS & PSO &FBA   \\ \midrule
        Emotion &100.7 &  162.4 & \underline{21.7}& 414.0 & 707.9  & 130.5 & \textbf{0.7} & \textit{73.8} & 120.2  \\ \midrule
        IMDB & \textit{155.7} &  431.4 & \underline{81.1} & 781.0 & 1007.9 & 170.5 & \textbf{3.7} & 166.9 & 159.2  \\ 
        \bottomrule
    \end{tabular}
\label{tab: efficiency}
\end{table*}
\subsection{Transferability}

The transferability of adversarial examples refers to the ability of an adversarial sample generated to deceive one model to deceive another model. This property is widely used as an important metric to evaluate the effectiveness of adversarial attacks. To assess the transferability of our method, we generate adversarial examples on BERT-C and TextCNN models and use them to attack the other model. The results of this evaluation are shown in Figure \ref{fig: transfer}, where the lower the classification accuracy, the higher the transferability. We observe that our method achieves the best transfer attack performance compared to other methods, indicating its ability to generate adversarial examples that can deceive different models. This property is particularly relevant in real-world scenarios where an attacker may not know the exact model deployed by the target and needs to generate adversarial examples that are effective against a range of models.

\subsection{Targeted Attacks}

A targeted adversarial attack aims to deceive the victim classifier by modifying the input sample in such a way that it is misclassified as a specified target class \(y'\) instead of its true class \(y\). Compared to untargeted attacks, targeted attacks are considered more harmful as they provide the attacker with more control over the final predicted label of the perturbed text. To conduct targeted attacks using FBA, we modify the definition of \(R\) in Eq. \ref{eqt: performance} by replacing \(1-F_{y}(x')\) with \(F_{y'}(x')\). To evaluate the effectiveness of FBA for targeted attacks, we perform experiments on the Emotion dataset and report the results in Table \ref{tab: target}. The results demonstrate that our proposed method achieves superior performance in terms of SAR, imperceptibility metrics (ROUGE, USE) and sentence quality metrics (GErr, PPL) for targeted attacks. This suggests that FBA can be effectively adapted for targeted adversarial attacks.

\subsection{Adversarial Retraining}
Adversarial retraining has emerged as a promising technique to improve the accuracy and robustness of machine learning models against adversarial attacks \cite{xi2022,ding2022,chivukula2020game}. By incorporating adversarial examples generated by attack algorithms during the training process, models can learn to handle adversarial perturbations better and improve their overall performance. Adversarial examples can be seen as informative features that reveal the weaknesses of the model, allowing it to learn more robust and discriminative features \cite{ilyas2019adversarial}. In order to assess the accuracy of retrained classifiers using adversarial examples, we added {1000, 2000, 3000, 4000, 5000, and 6000} randomly generated adversarial examples from the SST2 training set to the TextCNN model's training set. The accuracy of the model on the clean test set was then compared after adversarial training and after appending varying numbers of adversarial examples. Based on the results presented in Figure \ref{fig: adv_training}, we observed that the classifier trained using adversarial examples performed the best when the same number of adversarial examples were added from FBA.

\par

\subsection{Complexity and Qualitative Results}
Experiments were run on a RHEL 7.9 system with an Intel(R) Xeon(R) Gold 6238R CPU (2.2GHz, 28 cores - 26 enabled, 38.5MB L3 Cache), an NVIDIA Quadro RTX 5000 GPU (3072 Cores, 384 Tensor Cores, 16GB memory), and 88GB RAM.

Table \ref{tab: efficiency} presents the time taken to attack BERT and TextCNN classifiers on the Emotion dataset. Time efficiency is measured in seconds per example, where a lower value denotes better efficiency. As observed from Table \ref{tab: efficiency}, while our WMP and FBA methods take longer than certain static baselines like PWWS and BAE, they outperform others such as CLARE, FAGA, A2T, and BA in terms of efficiency. It is noted that the extended run time of our methods compared to some baseline approaches suggests the additional time invested in seeking more optimal adversarial examples.

\section{Conclusion and Future Work}\label{conclusion}
The FBA algorithm uses the Word Manipulation Process to create many possible textual perturbations that can fool a classifier, making it effective for targeted attacks. Adversarial retraining can defend against attacks but is expensive and may reduce model accuracy. Current defence methods for text are not sufficient, so developing better defences is a promising future direction.
\bibliographystyle{IEEEtran}
\bibliography{IEEEabrv,ref}

\begin{thebibliography}{10}
\providecommand{\url}[1]{#1}
\csname url@samestyle\endcsname
\providecommand{\newblock}{\relax}
\providecommand{\bibinfo}[2]{#2}
\providecommand{\BIBentrySTDinterwordspacing}{\spaceskip=0pt\relax}
\providecommand{\BIBentryALTinterwordstretchfactor}{4}
\providecommand{\BIBentryALTinterwordspacing}{\spaceskip=\fontdimen2\font plus
\BIBentryALTinterwordstretchfactor\fontdimen3\font minus
  \fontdimen4\font\relax}
\providecommand{\BIBforeignlanguage}[2]{{%
\expandafter\ifx\csname l@#1\endcsname\relax
\typeout{** WARNING: IEEEtran.bst: No hyphenation pattern has been}%
\typeout{** loaded for the language `#1'. Using the pattern for}%
\typeout{** the default language instead.}%
\else
\language=\csname l@#1\endcsname
\fi
#2}}
\providecommand{\BIBdecl}{\relax}
\BIBdecl

\bibitem{yang2021bigram}
X.~Yang, W.~Liu, J.~Bailey, D.~Tao, and W.~Liu, ``Bigram and unigram based text
  attack via adaptive monotonic heuristic search,'' in \emph{Proceedings of the
  AAAI Conference on Artificial Intelligence}, vol.~35, no.~1, 2021, pp.
  706--714.

\bibitem{wang2019towards}
W.~Wang, R.~Wang, L.~Wang, Z.~Wang, and A.~Ye, ``Towards a robust deep neural
  network against adversarial texts: A survey,'' \emph{IEEE Transactions on
  Knowledge and Data Engineering(TKDE)}, vol.~35, no.~3, pp. 3159--3179, 2023.

\bibitem{ebrahimi2017hotflip}
J.~Ebrahimi, A.~Rao, D.~Lowd, and D.~Dou, ``Hotflip: White-box adversarial
  examples for text classification,'' \emph{arXiv preprint arXiv:1712.06751},
  2017.

\bibitem{gan2019improving}
W.~C. Gan and H.~T. Ng, ``Improving the robustness of question answering
  systems to question paraphrasing,'' in \emph{Proceedings of the 57th Annual
  Meeting of the Association for Computational Linguistics}, 2019, pp.
  6065--6075.

\bibitem{hun2021gan}
H.~Sun, T.~Zhu, Z.~Zhang, D.~Jin, P.~Xiong, and W.~Zhou, ``Adversarial attacks
  against deep generative models on data: A survey,'' \emph{IEEE Transactions
  on Knowledge Data Engineering (TKDE)}, pp. 1--1, nov 2021.

\bibitem{garg2020bae}
S.~Garg and G.~Ramakrishnan, ``Bae: Bert-based adversarial examples for text
  classification,'' in \emph{Proceedings of the 2020 Conference on Empirical
  Methods in Natural Language Processing (EMNLP)}, 2020, pp. 6174--6181.

\bibitem{zang2020word}
Y.~Zang, F.~Qi, C.~Yang, Z.~Liu, M.~Zhang, Q.~Liu, and M.~Sun, ``Word-level
  textual adversarial attacking as combinatorial optimization,'' in
  \emph{Proceedings of the 58th Annual Meeting of the Association for
  Computational Linguistics}, 2020, pp. 6066--6080.

\bibitem{li2020bertattack}
L.~Li, R.~Ma, Q.~Guo, X.~Xue, and X.~Qiu, ``Bert-attack: Adversarial attack
  against bert using bert,'' in \emph{Proceedings of the 2020 Conference on
  Empirical Methods in Natural Language Processing (EMNLP)}, 2020, pp.
  6193--6202.

\bibitem{ni2022attacking}
M.~Ni, C.~Wang, T.~Zhu, S.~Yu, and W.~Liu, ``Attacking neural machine
  translations via hybrid attention learning,'' \emph{Machine Learning}, pp.
  1--26, 2022.

\bibitem{ren2019pwws}
S.~Ren, Y.~Deng, K.~He, and W.~Che, ``Generating natural language adversarial
  examples through probability weighted word saliency,'' in \emph{Proceedings
  of the 57th annual meeting of the association for computational linguistics},
  2019, pp. 1085--1097.

\bibitem{Goodfellow2015fgsm}
I.~J. Goodfellow, J.~Shlens, and C.~Szegedy, ``Explaining and harnessing
  adversarial examples,'' \emph{CoRR}, vol. abs/1412.6572, 2015.

\bibitem{PapernotMSH16}
N.~Papernot, P.~D. McDaniel, A.~Swami, and R.~E. Harang, ``Crafting adversarial
  input sequences for recurrent neural networks,'' \emph{CoRR}, vol.
  abs/1604.08275, 2016.

\bibitem{alzantot2018generating}
M.~Alzantot, Y.~Sharma, A.~Elgohary, B.-J. Ho, M.~Srivastava, and K.-W. Chang,
  ``Generating natural language adversarial examples,'' in \emph{Proceedings of
  the 2018 Conference on Empirical Methods in Natural Language Processing},
  2018, pp. 2890--2896.

\bibitem{jia2019faga}
R.~Jia, A.~Raghunathan, K.~G{\"o}ksel, and P.~Liang, ``Certified robustness to
  adversarial word substitutions,'' in \emph{Proceedings of the 2019 Conference
  on Empirical Methods in Natural Language Processing and the 9th International
  Joint Conference on Natural Language Processing (EMNLP-IJCNLP)}, 2019, pp.
  4129--4142.

\bibitem{pennington2014glove}
J.~Pennington, R.~Socher, and C.~D. Manning, ``Glove: Global vectors for word
  representation,'' in \emph{Empirical Methods in Natural Language Processing
  (EMNLP)}, 2014, pp. 1532--1543.

\bibitem{Mikolov2013word2vect}
T.~Mikolov, K.~Chen, G.~S. Corrado, and J.~Dean, ``Efficient estimation of word
  representations in vector space,'' in \emph{ICLR}, 2013.

\bibitem{mrkvsic2016counter}
N.~Mrk{\v{s}}i{\'c}, D.~{\'O}. S{\'e}aghdha, B.~Thomson, M.~Gasic, L.~M.~R.
  Barahona, P.-H. Su, D.~Vandyke, T.-H. Wen, and S.~Young, ``Counter-fitting
  word vectors to linguistic constraints,'' in \emph{Proceedings of the 2016
  Conference of the North American Chapter of the Association for Computational
  Linguistics: Human Language Technologies}, 2016, pp. 142--148.

\bibitem{miller1990wordnet}
G.~A. Miller, R.~Beckwith, C.~Fellbaum, D.~Gross, and K.~J. Miller,
  ``Introduction to wordnet: An on-line lexical database,'' \emph{International
  journal of lexicography}, vol.~3, no.~4, pp. 235--244, 1990.

\bibitem{dong2003hownet}
Z.~Dong and Q.~Dong, ``Hownet-a hybrid language and knowledge resource,'' in
  \emph{International Conference on Natural Language Processing and Knowledge
  Engineering, 2003. Proceedings. 2003}.\hskip 1em plus 0.5em minus 0.4em\relax
  IEEE, 2003, pp. 820--824.

\bibitem{Devlin2019BERTPO}
J.~Devlin, M.-W. Chang, K.~Lee, and K.~Toutanova, ``Bert: Pre-training of deep
  bidirectional transformers for language understanding,'' in \emph{NAACL},
  2019.

\bibitem{Liu2019RoBERTaAR}
Y.~Liu, M.~Ott, N.~Goyal, J.~Du, M.~Joshi, D.~Chen, O.~Levy, M.~Lewis,
  L.~Zettlemoyer, and V.~Stoyanov, ``Roberta: A robustly optimized bert
  pretraining approach,'' \emph{ArXiv}, vol. abs/1907.11692, 2019.

\bibitem{rubinstein1999crossoptimization}
R.~Rubinstein, ``The cross-entropy method for combinatorial and continuous
  optimization,'' \emph{Methodology and computing in applied probability},
  vol.~1, no.~2, pp. 127--190, 1999.

\bibitem{fan2018rectangular}
X.~Fan, B.~Li, and S.~Sisson, ``Rectangular bounding process,'' \emph{Advances
  in Neural Information Processing Systems}, vol.~31, 2018.

\bibitem{haase2021quantum}
J.~F. Haase, L.~Dellantonio, A.~Celi, D.~Paulson, A.~Kan, K.~Jansen, and C.~A.
  Muschik, ``A resource efficient approach for quantum and classical
  simulations of gauge theories in particle physics,'' \emph{Quantum}, vol.~5,
  p. 393, 2021.

\bibitem{metropolis1953MH}
N.~Metropolis, A.~W. Rosenbluth, M.~N. Rosenbluth, A.~H. Teller, and E.~Teller,
  ``Equation of state calculations by fast computing machines,'' \emph{The
  journal of chemical physics}, vol.~21, no.~6, pp. 1087--1092, 1953.

\bibitem{hastings1970monte}
\BIBentryALTinterwordspacing
W.~K. Hastings, ``Monte carlo sampling methods using markov chains and their
  applications,'' \emph{Biometrika}, vol.~57, pp. 97--109, 1970. [Online].
  Available: \url{https://api.semanticscholar.org/CorpusID:21204149}
\BIBentrySTDinterwordspacing

\bibitem{Kumagai2016HumanlikeNL}
K.~Kumagai, I.~Kobayashi, D.~Mochihashi, H.~Asoh, T.~Nakamura, and T.~Nagai,
  ``Human-like natural language generation using monte carlo tree search,'' in
  \emph{CC-NLG}, 2016.

\bibitem{harrison2017toward}
B.~Harrison, C.~Purdy, and M.~O. Riedl, ``Toward automated story generation
  with markov chain monte carlo methods and deep neural networks,'' in
  \emph{Thirteenth Artificial Intelligence and Interactive Digital
  Entertainment Conference}, 2017.

\bibitem{Kang2011SamplingLE}
X.~Kang and F.~Ren, ``Sampling latent emotions and topics in a hierarchical
  bayesian network,'' \emph{2011 7th International Conference on Natural
  Language Processing and Knowledge Engineering}, pp. 37--42, 2011.

\bibitem{Singh2012MonteCM}
S.~Singh, M.~L. Wick, and A.~McCallum, ``Monte carlo mcmc: Efficient inference
  by sampling factors,'' in \emph{AKBC-WEKEX@NAACL-HLT}, 2012.

\bibitem{Conneau2020UnsupervisedCR}
A.~Conneau, K.~Khandelwal, N.~Goyal, V.~Chaudhary, G.~Wenzek, F.~Guzm{\'a}n,
  E.~Grave, M.~Ott, L.~Zettlemoyer, and V.~Stoyanov, ``Unsupervised
  cross-lingual representation learning at scale,'' in \emph{ACL}, 2020.

\bibitem{song2020mpnet}
K.~Song, X.~Tan, T.~Qin, J.~Lu, and T.-Y. Liu, ``Mpnet: Masked and permuted
  pre-training for language understanding,'' \emph{Advances in Neural
  Information Processing Systems}, vol.~33, pp. 16\,857--16\,867, 2020.

\bibitem{li2021clare}
D.~Li, Y.~Zhang, H.~Peng, L.~Chen, C.~Brockett, M.-T. Sun, and W.~B. Dolan,
  ``Contextualized perturbation for textual adversarial attack,'' in
  \emph{Proceedings of the 2021 Conference of the North American Chapter of the
  Association for Computational Linguistics: Human Language Technologies},
  2021, pp. 5053--5069.

\bibitem{Jin2020IsBR}
D.~Jin, Z.~Jin, J.~T. Zhou, and P.~Szolovits, ``Is bert really robust? a strong
  baseline for natural language attack on text classification and entailment,''
  in \emph{AAAI}, 2020.

\bibitem{Wang2019NaturalLA}
\BIBentryALTinterwordspacing
X.~Wang, H.~Jin, and X.~Wang, ``Natural language adversarial attacks and
  defenses in word level,'' \emph{ArXiv}, vol. abs/1909.06723, 2019. [Online].
  Available: \url{https://api.semanticscholar.org/CorpusID:202577228}
\BIBentrySTDinterwordspacing

\bibitem{pavliotis2015stochastic}
G.~A. Pavliotis, ``Stochastic processes and applications,'' \emph{Informe
  t{\'e}cnico}, 2015.

\bibitem{kroesehandbook}
D.~P. Kroese, T.~Taimre, and Z.~I. Botev, \emph{Handbook of Monte Carlo
  Methods}.\hskip 1em plus 0.5em minus 0.4em\relax John Wiley \& Sons, 2011.

\bibitem{cer2018universal}
D.~Cer, Y.~Yang, S.-y. Kong, N.~Hua, N.~Limtiaco, R.~S. John, N.~Constant,
  M.~Guajardo-Cespedes, S.~Yuan, C.~Tar \emph{et~al.}, ``Universal sentence
  encoder for english,'' in \emph{Proceedings of the 2018 conference on
  empirical methods in natural language processing: system demonstrations},
  2018, pp. 169--174.

\bibitem{ag_news}
X.~Zhang, J.~J. Zhao, and Y.~LeCun, ``Character-level convolutional networks
  for text classification,'' in \emph{NIPS}, 2015.

\bibitem{emotion}
E.~Saravia, H.-C.~T. Liu, Y.-H. Huang, J.~Wu, and Y.-S. Chen, ``{CARER}:
  Contextualized affect representations for emotion recognition,'' in
  \emph{Proceedings of the 2018 Conference on Empirical Methods in Natural
  Language Processing}.\hskip 1em plus 0.5em minus 0.4em\relax Brussels,
  Belgium: Association for Computational Linguistics, Oct.-Nov. 2018, pp.
  3687--3697.

\bibitem{sst2}
R.~Socher, A.~Perelygin, J.~Wu, J.~Chuang, C.~D. Manning, A.~Ng, and C.~Potts,
  ``Recursive deep models for semantic compositionality over a sentiment
  treebank,'' in \emph{Proceedings of the 2013 Conference on Empirical Methods
  in Natural Language Processing}.\hskip 1em plus 0.5em minus 0.4em\relax
  Seattle, Washington, USA: Association for Computational Linguistics, Oct.
  2013, pp. 1631--1642.

\bibitem{mingze2023}
M.~Ni, Z.~Sun, and W.~Liu, ``Fraud’s bargain attacks to textual classifiers
  via metropolis-hasting sampling (student abstract),'' in \emph{Proceedings of
  the AAAI Conference on Artificial Intelligence}, 2023.

\bibitem{yoo2021a2t}
J.~Y. Yoo and Y.~Qi, ``Towards improving adversarial training of nlp models,''
  in \emph{Findings of the Association for Computational Linguistics: EMNLP
  2021}, 2021, pp. 945--956.

\bibitem{naber2003rule}
D.~Naber \emph{et~al.}, ``A rule-based style and grammar checker,''
  \emph{Citeseer}, 2003.

\bibitem{tradeoff}
P.~Michel, X.~Li, G.~Neubig, and J.~M. Pino, ``On evaluation of adversarial
  perturbations for sequence-to-sequence models,'' \emph{arXiv preprint
  arXiv:1903.06620}, 2019.

\bibitem{xi2022}
H.~Xia, S.~Shao, C.~Hu, R.~Zhang, T.~Qiu, and F.~Xiao, ``Robust clustering
  model based on attention mechanism and graph convolutional network,''
  \emph{IEEE Transactions on Knowledge and Data Engineering (TKDE)}, pp. 1--1,
  2022.

\bibitem{ding2022}
X.~Ding, H.~Fang, Z.~Zhang, K.-K.~R. Choo, and H.~Jin, ``Privacy-preserving
  feature extraction via adversarial training,'' \emph{IEEE Transactions on
  Knowledge and Data Engineering (TKDE)}, vol.~34, no.~4, pp. 1967--1979, 2022.

\bibitem{chivukula2020game}
A.~S. Chivukula, X.~Yang, W.~Liu, T.~Zhu, and W.~Zhou, ``Game theoretical
  adversarial deep learning with variational adversaries,'' \emph{IEEE
  Transactions on Knowledge and Data Engineering (TKDE)}, vol.~33, no.~11, pp.
  3568--3581, 2020.

\bibitem{ilyas2019adversarial}
A.~Ilyas, S.~Santurkar, D.~Tsipras, L.~Engstrom, B.~Tran, and A.~Madry,
  ``Adversarial examples are not bugs, they are features,'' \emph{Advances in
  neural information processing systemas}, vol.~32, 2019.

\end{thebibliography}

\begin{IEEEbiography}[{\includegraphics[width=1in,height=1.25in,clip,keepaspectratio]{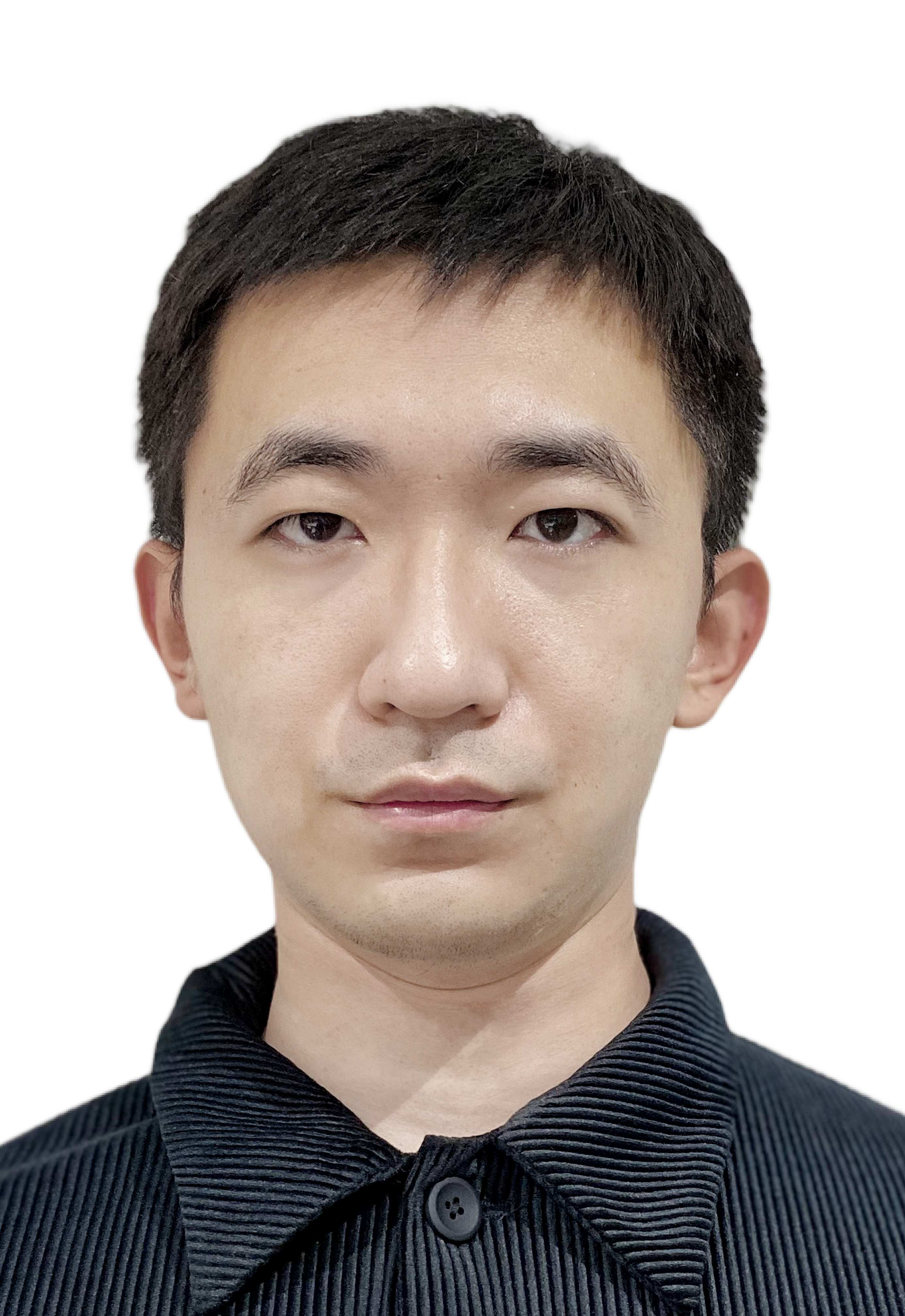}}]{Mingze Ni}
graduated from Australian National University with bachelor’s degrees in science (Statistics and Mathematics) and the University of Queensland with a Bachelor of Statistics (Honours). He is currently a PhD student at the University of Technology Sydney. His researches focus on adversarial machine learning, textual attacks and natural language processing.
\end{IEEEbiography}

\begin{IEEEbiography}[{\includegraphics[width=1in,height=1.25in,clip,keepaspectratio]{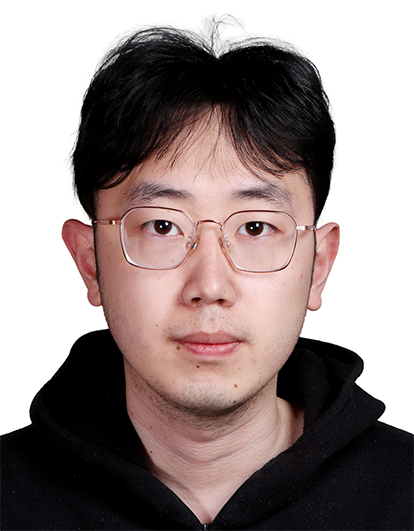}}]{Zhensu Sun}
graduated from Tongji University, China, with a master's degree in software engineering. Currently, he is a research assistant at ShanghaiTech University, China. His research interest is the application of deep learning in software engineering to boost developers' productivity.
\end{IEEEbiography}

%\newpage

\begin{IEEEbiography}[{\includegraphics[width=1in,height=1.25in,clip,keepaspectratio]{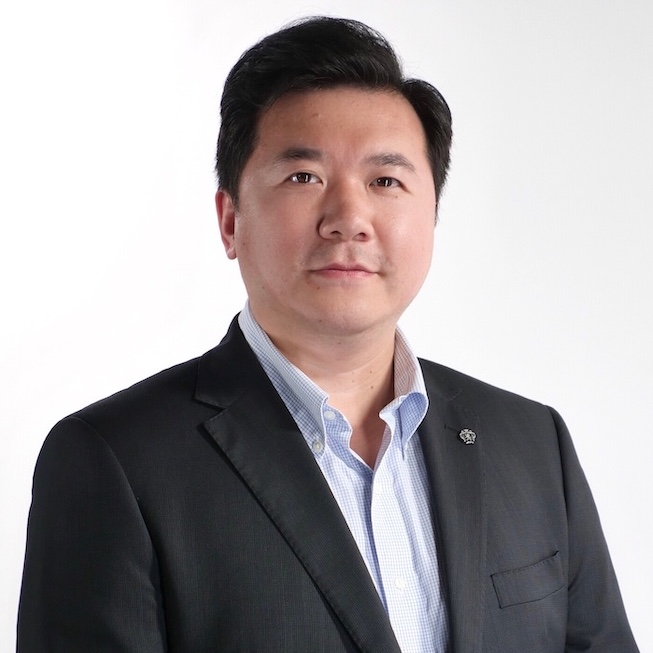}}]{Wei Liu} (M’15-SM’20) received the PhD degree in machine learning research from the University of Sydney in 2011. He is currently the Director of Future Intelligence Research Lab, and an Associate Professor in Machine Learning, at the School of Computer Science, the University of Technology Sydney (UTS), Australia. Before joining UTS, he was a Research Fellow at the University of Melbourne and then a Machine Learning Researcher at NICTA. His current research focuses are adversarial machine learning, game theory, causal inference, multimodal learning, and natural language processing. He has published more than 100 papers in top-tier journals and conferences. Additionally, he has won three Best Paper Awards and one Most Influential Paper Award.
\end{IEEEbiography}


%\vfill


\end{document}